\documentclass{article}

 \usepackage[preprint]{neurips_2026}

\usepackage[utf8]{inputenc} %
\usepackage[T1]{fontenc}    %
\usepackage{hyperref}       %
\usepackage{url}            %
\usepackage{booktabs}       %
\usepackage{amsfonts}       %
\usepackage{nicefrac}       %
\usepackage{microtype}      %
\usepackage{xcolor}         %
\usepackage{graphicx}
\usepackage{wrapfig}
\input{insbox}
\usepackage{multirow}
\usepackage{caption} %
\title{PresentAgent-2: Towards Generalist Multimodal Presentation Agents}

\author{
Wei Wu$^{1*}$\quad
Ziyang Xu$^{1*}$\quad
Zeyu Zhang$^{1*\dag}$\quad
Yang Zhao$^{2}$\quad
Hao Tang$^{1\ddag}$\\ 
[0.3em]
$^1$Peking University\quad
$^2$La Trobe University\\
[0.1em]
\footnotesize $^*$Equal contribution.
$^\dag$Project lead.
$^\ddag$Corresponding author: bjdxtanghao@gmail.com.
}

\begin{document}

\maketitle

\begin{abstract}
  Presentation generation is moving beyond static slide creation toward end-to-end presentation video generation with research grounding, multimodal media, and interactive delivery. We introduce PresentAgent-2, an agentic framework for generating presentation videos from user queries. Given an open-ended user query and a selected presentation mode, PresentAgent-2 first summarizes the query into a focused topic and performs deep research over presentation-friendly sources to collect multimodal resources, including relevant text, images, GIFs, and videos. It then constructs presentation slides, generates mode-specific scripts, and composes slides, audio, and dynamic media into a complete presentation video. PresentAgent-2 supports three independent presentation modes within a unified framework: Single Presentation, which generates a single-speaker narrated presentation video; Discussion, which creates a multi-speaker presentation with structured speaker roles, such as for asking guiding questions, explaining concepts, clarifying details, and summarizing key points; and Interaction, which independently supports answering audience questions grounded in the generated slides, scripts, retrieved evidence, and presentation context. To evaluate these capabilities, we build a multimodal presentation benchmark covering single presentation, discussion, and interaction scenarios, with task-specific evaluation criteria for content quality, media relevance, dynamic media use, dialogue naturalness, and interaction grounding. Overall, PresentAgent-2 extends presentation generation from document-dependent slide creation to query-driven, research-grounded presentation video generation with multimodal media, dialogue, and interaction.
  Code:~\url{https://github.com/AIGeeksGroup/PresentAgent-2}.
  Website:~\url{https://aigeeksgroup.github.io/PresentAgent-2}.
\end{abstract}

\section{Introduction}
Presentation videos are an important medium for communicating knowledge. They combine structured slides, spoken explanations, and visual examples, making complex topics easier to follow than static documents or slide images alone. In education, research communication, and technical explanation, a good presentation video does not merely summarize content; it organizes information into a clear structure, highlights important visual evidence, and delivers the material in a form that an audience can understand.
\begin{figure*}[t]
\centering
\setlength{\tabcolsep}{3pt}
\setlength{\fboxsep}{0pt}
\setlength{\fboxrule}{0.35pt}

\newcommand{\teaserfig}[1]{%
\fbox{%
    \includegraphics[width=0.32\textwidth]{#1}%
}
}

\begin{tabular}{@{}ccc@{}}
\teaserfig{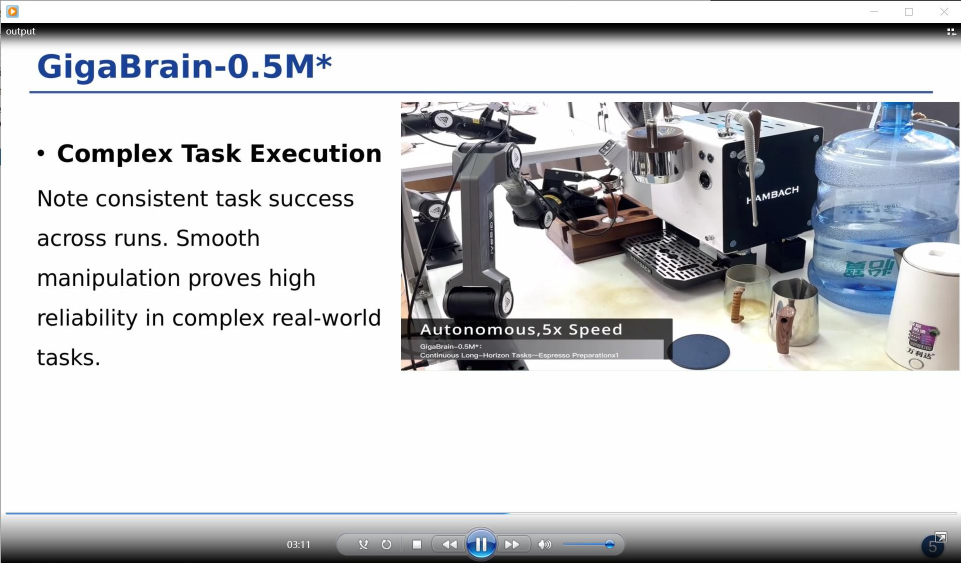}
&
\teaserfig{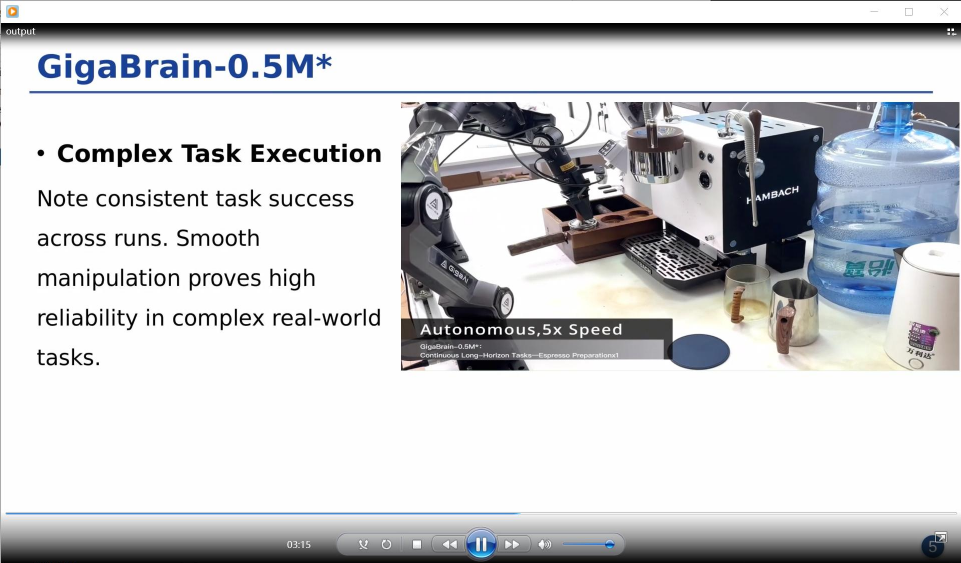}
&
\teaserfig{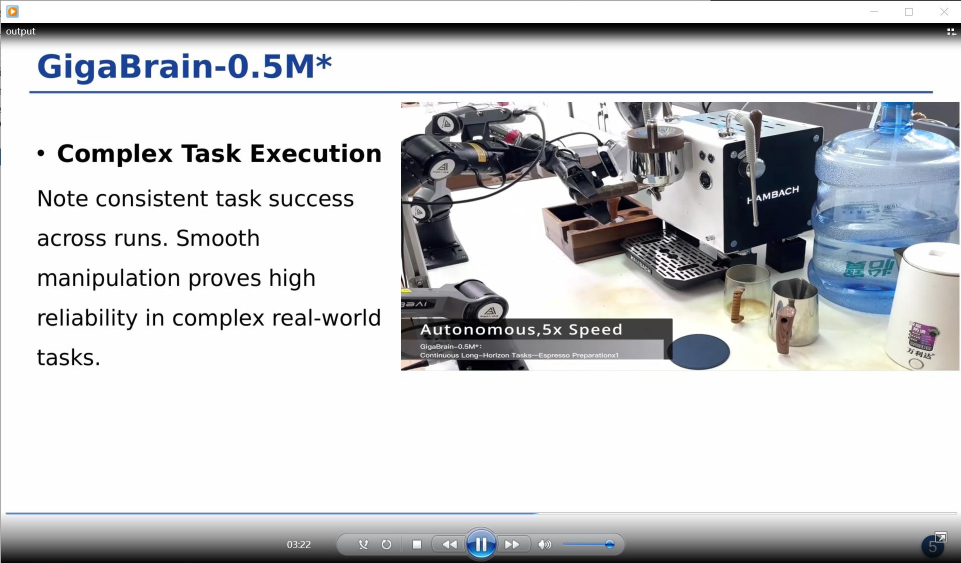}
\end{tabular}

\caption{
\textbf{Representative frames from a generated PresentAgent-2 presentation video.}
The frames are sampled from different timestamps of the same video, showing how retrieved video evidence is incorporated into the generated presentation.
}
\label{fig:teaser_coffee}
\end{figure*}
Recent work has made substantial progress in automatically generating research communication materials. Paper2Poster \cite{pang2025paper2poster} studies how to compress scientific papers into visually coherent posters. PresentAgent \cite{shi2025presentagent} extends document-to-slide generation toward narrated presentation videos from long-form documents. Paper2Video and VideoAgent \cite{zhu2025paper2video, liang2025videoagent} further study academic presentation video generation from research papers, integrating slides, subtitles, speech, cursor grounding, and talking-head rendering. These works show that LLM- and VLM-based agents can organize long documents, design visual layouts, synthesize narration, and evaluate whether the generated results effectively convey knowledge.

However, these methods mostly assume that the source content is already given as a complete document, such as a paper, report, or technical blog \cite{jung2025talk, zheng2025pptagent, yang2025auto}. They focus on converting existing content into a visual or presentation output, rather than generating a presentation video from a short and open-ended user query. This assumption limits their applicability in many practical scenarios. A user may simply ask, ``Please explain flow matching'', without providing a paper or report. In this setting, the system must first determine what should be explained, retrieve reliable supporting materials, select suitable visual and dynamic media, and then construct a coherent presentation video \cite{kyaw2025node, hu2025polyvivid, kong2025let}.

We therefore study query-to-presentation video generation. Given a natural-language query, the goal is to generate a presentation-style video that explains the requested topic. This task is challenging because the input query does not contain the full content or visual resources needed for slide construction, while the output should still be a structured presentation video.

\begin{figure*}[h]
  \centering
  \includegraphics[width=\textwidth]{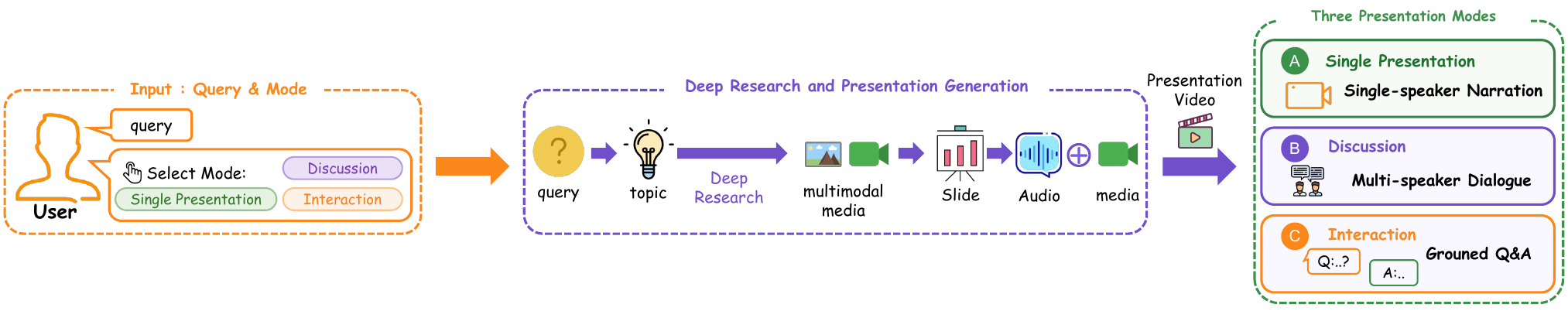}
  \caption{
    \textbf{Overview of PresentAgent-2.}
    PresentAgent-2 turns a user query into a presentation video through deep research, slide/script generation, audio synthesis, and video composition.
  }
  \label{fig:teaser}
\end{figure*}

To tackle these challenges, we propose PresentAgent-2, an agentic framework for query-driven presentation video generation, as illustrated in Figure~\ref{fig:teaser}. Given a user query, the system first summarizes it into a focused topic and performs deep research to search for candidate sources, such as webpages, tutorials, demo pages, and articles with clear explanations or visual examples. It then filters these sources and extracts a multimodal resource set, including textual content, images, GIFs, and videos. Based on the retrieved resources, PresentAgent-2 plans the presentation structure, generates slides and scripts, converts scripts into audio, and composes the slides, audio, and media into the final presentation video. Importantly, for GIFs and videos, PresentAgent-2 does not turn them into static screenshots. Instead, during video composition, it places each dynamic medium in the corresponding slide region, so that videos, animations, and moving examples can keep playing inside PPT-style pages.

PresentAgent-2 supports three independent presentation video modes within a unified framework. Single Presentation generates a single-speaker video that explains the content following the slide order. Discussion generates a multi-speaker dialogue, in which different speakers take different roles, such as asking guiding questions, explaining concepts, clarifying details, and summarizing key points. Interaction supports an interactive presentation format, in which the system answers audience questions grounded in the generated slides, scripts, retrieved evidence, and presentation context. These three modes share the same deep research and presentation generation backbone, but differ in their script structure and delivery style.

We further build a multimodal presentation benchmark for evaluating query-driven presentation videos across three scenarios: single presentation, discussion presentation, and interactive presentation. The benchmark evaluates general presentation quality, multimodal media use, discussion quality, and interaction grounding. This benchmark reflects the central challenge of our task: a generated presentation video should not only be factually correct, but also communicate knowledge through structured slides, appropriate media, and mode-specific delivery.
Our contributions are summarized as follows:
\begin{itemize}
    \item We propose PresentAgent-2, a query-driven presentation video generation framework that integrates topic understanding, deep research, multimodal resource retrieval, slide-and-script generation, and video composition. Starting from an open-ended user query, the system actively collects textual and multimodal resources, including images, GIFs, and videos, and composes them into structured presentation videos while preserving dynamic media.

    \item We support three independent presentation video modes within a unified framework: Single Presentation, Discussion, and Interaction. These modes correspond to single-speaker narration, multi-speaker dialogue, and grounded interactive Q\&A, enabling different forms of presentation delivery from the same researched content.

    \item We build a multimodal presentation benchmark for evaluating query-driven presentation videos across single presentation, discussion, and interaction scenarios, covering general presentation quality, multimodal media use, discussion quality, and interaction grounding.
\end{itemize}

\section{Related Work}

\subsection{Presentation Generation from Documents}

Early work on automated presentation creation mainly frames the task as multimodal document summarization, involving document understanding, content abstraction, and visual layout prediction~\cite{ge2025autopresent, wang2025infinity}. Representative systems such as Doc2PPT establish evaluation criteria for slide quality, while SlideGen and Paper2Poster further improve slide or poster generation through multimodal agents and layout-aware visual organization~\cite{fu2022doc2ppt, konstantinov2026slides, liang2025slidegen, pang2025paper2poster}. However, these methods largely treat presentations as static content carriers: they generate visual layouts from given documents but do not address oral delivery, dynamic media composition, or open-ended user queries~\cite{liu2025presenting}. Tool-augmented and multimodal reasoning frameworks further enable language models to invoke visual tools and process multimodal inputs~\cite{yang2023gpt4tools, yang2023mm}, but they lack presentation-specific constraints for coordinating slides, scripts, audio, and rhetorical structures such as guiding questions, conceptual explanations, and summaries~\cite{sun2025genesis}.

\subsection{Presentation Video and Multimodal Content Synthesis}

General multimodal generation models provide useful components for presentation synthesis, including video generation, speech generation, temporal alignment, motion generation, long-sequence modeling, and multimodal evaluation~\cite{li2023videogen, xue2025phyt2v, yang2024cogvideox, zhao2025unified, team2026qwen3, zhang2025motion, zhang2024infinimotion, zhang2024kmm, zhang2024motion, li2023evaluating}. Interactive visual instruction models also support multimodal instruction following and visual question answering~\cite{wu2025towards}. However, these techniques are usually evaluated as standalone generation or understanding modules, and have not been integrated into a complete presentation workflow with research-based retrieval, slide-level planning, structured script writing, dynamic media composition, and interactive delivery~\cite{wang2026mavis}.

Recent studies move closer to end-to-end presentation video generation~\cite{hu2025multimodal}. PresentAgent converts long documents into narrated presentation videos by coordinating slide assembly, script generation, and audio-visual synchronization~\cite{shi2025presentagent}. Paper2Video and VideoAgent generate scientific explanation videos from academic papers with subtitles, narration, and animation rendering~\cite{zhu2025paper2video, liang2025videoagent}. Other agent-based systems improve presentation or multimodal content creation through visual self-correction, presentation coaching, and prompt-based iterative refinement~\cite{xu2025pregenie, chen2025presentcoach, kyaw2025node}. Despite this progress, existing systems still primarily rely on provided source documents or focus on single-speaker and paper-specific scenarios. They do not unify query-driven research retrieval, multi-speaker dialogue simulation, structured role setting, dynamic media use, and grounded audience interaction within one presentation generation framework~\cite{deng2025emerging, xie2024show, lin2025showui}.

\section{PresentEval: A Multimodal Presentation Benchmark}
\label{sec:benchmark}
The benchmark supports the evaluation of query-to-presentation video generation across three independent presentation modes: Single Presentation, Discussion, and Interaction. Different from document-to-presentation benchmarks that generate from a given source document, our benchmark uses open-ended user queries as input. Each benchmark example contains a query and a human-created reference presentation video, while the system is only given the query during generation. This setting evaluates whether a system can recover missing context through deep research, organize the information into a structured presentation, and generate a presentation video in the specified mode.

\subsection{Dataset Construction}

\paragraph{Data Source.}
We collect 60 high-quality query--reference video pairs to construct the multimodal presentation benchmark. The reference videos are collected from public video platforms, educational repositories, and professional presentation archives. Each reference video follows a presentation-style format and communicates knowledge through slides, speech, visual examples, discussion, or audience interaction. For each reference video, we formulate an open-ended user query that simulates what a real user might ask when requesting such a presentation. Unlike document-to-presentation benchmarks, we do not provide the source document, paper, or report used to create the reference video; the query alone serves as the system input.

\paragraph{Data Statistics.}
To evaluate different presentation modes, we organize the 60 examples into three independent mode-specific sets: Single Presentation, Discussion, and Interaction, with 20 examples in each set. The Single Presentation set contains 20 single-speaker narrated presentations for evaluating query-driven single-speaker presentation video generation. The Discussion set contains 20 multi-speaker presentation-style discussions for evaluating discussion-style presentation video generation. The Interaction set contains 20 presentations with audience questions or interactive explanations for evaluating interactive presentation and grounded question answering. These three sets correspond to different presentation modes, delivery formats, and evaluation focuses. All reference videos are approximately 5--7 minutes long, which is long enough to cover a complete presentation flow while remaining suitable for human evaluation and VLM-based evaluation.

\subsection{Evaluation Metrics}
\begin{wrapfigure}[19]{r}{0.50\linewidth}
  \vspace{-0.8em}
  \centering
  \includegraphics[width=\linewidth]{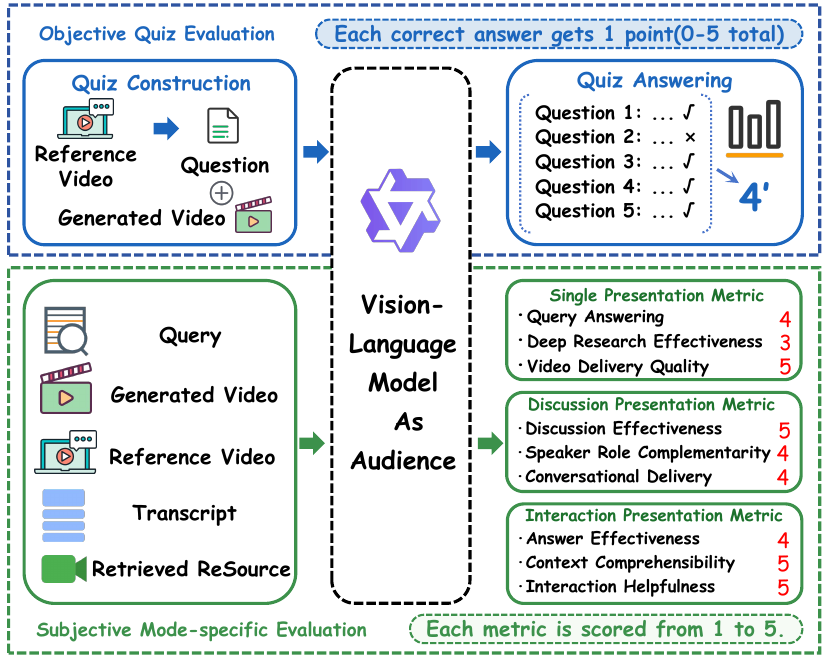}
  \caption{Evaluation pipeline. Objective quiz evaluation measures knowledge delivery, while subjective evaluation scores mode-specific presentation quality.
  }
  \label{fig:evaluation_pipeline}
  \vspace{-1.2em}
\end{wrapfigure}
As shown in Figure~\ref{fig:evaluation_pipeline}, we evaluate generated presentation videos using two components: objective quiz evaluation and subjective mode-specific evaluation. Objective quiz evaluation measures whether the generated video conveys the key knowledge required by the user query. Subjective mode-specific evaluation assesses whether the generated result satisfies the quality requirements of the selected presentation mode. Together, this design evaluates both audience comprehension and mode-specific presentation quality.

\paragraph{Objective Quiz Evaluation.}

Objective quiz evaluation consists of two stages: quiz construction and quiz answering. In the quiz construction stage, for each query--reference video pair, we construct five multiple-choice questions based on the reference presentation video and the expected knowledge points of the query. Each question contains four options with one correct answer, and the reference video is used to annotate the answer key. In the quiz answering stage, the VLM acts as an audience member and answers these questions using only the generated video and the transcript transcribed from the generated video's audio. Each correct answer receives one point, while an incorrect answer receives zero points; therefore, the quiz score ranges from 0 to 5. Each generated video receives one quiz score, and the reported quiz scores are averaged over all examples in the corresponding mode and model. This score measures how effectively the generated presentation communicates the requested knowledge. Table~\ref{tab:objective_quiz_examples} shows representative quiz examples, with correct answers highlighted in bold.

\begin{table*}[t]
  \caption{
  Example multiple-choice questions for objective quiz evaluation.
  Each question set is constructed from the corresponding reference presentation video and user query.
  Correct options are highlighted in bold.
  }
  \label{tab:objective_quiz_examples}
  \centering
  \footnotesize
  \setlength{\tabcolsep}{4pt}
  \renewcommand{\arraystretch}{1.15}
  \begin{tabular}{@{} p{0.35\textwidth} p{0.60\textwidth} @{}}
    \toprule
    \textbf{Mode and Question} & \textbf{Options} \\
    \midrule
    
    \textbf{Single Presentation} \newline
    What is the main idea of flow matching in generative modeling?
    &
    A. Learning a fixed dataset classifier \newline
    B. \textbf{Matching a continuous transformation path} \newline
    C. Compressing images into discrete tokens \newline
    D. Training without any learned dynamics \\
    
    \midrule
    
    \textbf{Discussion Presentation} \newline
    What key contrast distinguishes diffusion models from flow matching?
    &
    A. \textbf{Diffusion removes noise; flow matching learns a transformation path} \newline
    B. Both methods only classify images \newline
    C. Flow matching requires no training objective \newline
    D. Diffusion models cannot generate samples \\

    \midrule

    \textbf{Interaction Presentation} \newline
    When an audience member asks why flow matching can be more efficient than diffusion models, what is the best answer?
    &
    A. Flow matching avoids modeling data transformations. \newline
    B. Flow matching replaces generation with classification. \newline
    C. \textbf{Flow matching learns a continuous path and often needs fewer sampling steps.} \newline
    D. Flow matching only works for Gaussian distributions. \\
    
    \bottomrule
  \end{tabular}
\end{table*}

\begin{table*}[t]
  \centering
  \caption{
  Mode-specific subjective metrics. Each metric is scored independently on a 1--5 scale by the VLM judge. 
  Abbreviations: QA = Query Answering; DRE = Deep Research Effectiveness; VDQ = Video Delivery Quality; 
  DE = Discussion Effectiveness; SRC = Speaker Role Complementarity; CD = Conversational Delivery; 
  AE = Answer Effectiveness; CC = Content Comprehensibility; IH = Interaction Helpfulness.
  }
  \label{tab:subjective_metrics}
  \footnotesize
  \setlength{\tabcolsep}{5pt}
  \renewcommand{\arraystretch}{1.10}

  \begin{tabular}{@{} p{0.20\textwidth} p{0.08\textwidth} p{0.66\textwidth} @{}}
    \toprule
    \textbf{Mode} & \textbf{Metric} & \textbf{Criterion} \\
    \midrule

    \multirow[t]{3}{0.20\textwidth}{\vspace{0pt}\textbf{Single\\Presentation}}
    & QA
    & Directly answers the query and covers key topic concepts. \\
    & DRE
    & Uses relevant textual and multimodal resources to support the explanation. \\
    & VDQ
    & Delivers coherent content through slides, narration, and visuals. \\

    \midrule

    \multirow[t]{3}{0.20\textwidth}{\vspace{0pt}\textbf{Discussion\\Presentation}}
    & DE
    & Uses dialogue to clarify, compare, and extend the presented ideas. \\
    & SRC
    & Maintains complementary speaker roles for questioning, explaining, and summarizing. \\
    & CD
    & Provides natural, coherent, and easy-to-follow conversation. \\

    \midrule

    \multirow[t]{3}{0.20\textwidth}{\vspace{0pt}\textbf{Interaction\\Presentation}}
    & AE
    & Answers audience questions correctly and directly. \\
    & CC
    & Provides clear, understandable, and unambiguous answers. \\
    & IH
    & Offers useful clarification that supports audience understanding. \\

    \bottomrule
  \end{tabular}
\end{table*}
\paragraph{Subjective Mode-specific Evaluation.}
We further use the VLM as an audience member for subjective scoring to evaluate presentation quality. For each generated video, the VLM judge receives the user query, generated video, reference video, retrieved resources, and the transcript transcribed from the generated video's audio, and assigns independent 1--5 scores to the three metrics defined for the corresponding presentation mode. As shown in Table~\ref{tab:subjective_metrics}, Single Presentation evaluates query answering, deep research effectiveness, and video delivery quality; Discussion Presentation evaluates dialogue effectiveness, speaker role complementarity, and conversational delivery; and Interaction Presentation evaluates answer effectiveness, content comprehensibility, and interaction helpfulness. We provide additional evaluation prompts, scoring rules, and metric-specific rubrics in Appendix~\ref{sec:app_eval_prompts}.

\section{Method: PresentAgent-2}
\begin{figure*}[t]
  \centering
  \includegraphics[width=\textwidth]{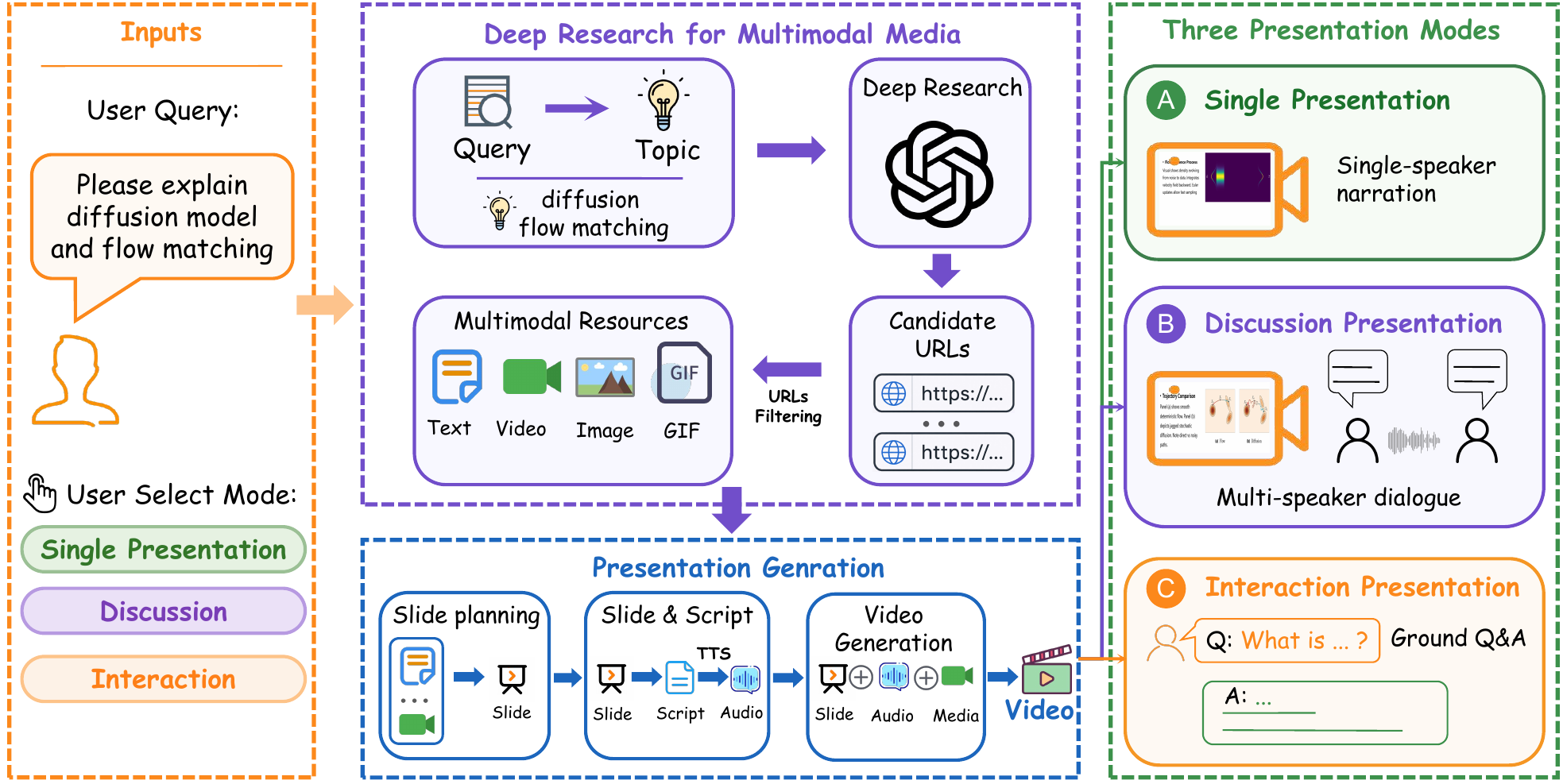}
  \caption{Overview of the PresentAgent-2 framework. Given a user query and a selected presentation mode, PresentAgent-2 first performs deep research to collect multimodal resources, then constructs presentation content, and finally generates a presentation video in single presentation, discussion, or interaction mode.}
  \label{fig:overview}
\end{figure*}

Existing presentation generation systems often assume that users provide a source document, such as a paper or report. To relax this requirement, we introduce PresentAgent-2, a multimodal agent that generates presentation videos from user queries. Given a query and a user-selected presentation mode, PresentAgent-2 first summarizes the query into a topic and performs deep research to collect topic-relevant text and multimodal media. It then uses these resources to construct presentation content. The system contains three core components: deep research for multimodal resources, a shared presentation generation backbone, and three supported presentation video modes: \emph{Single Presentation}, \emph{Discussion}, and \emph{Interaction}. Figure~\ref{fig:overview} shows the overall workflow of PresentAgent-2. The following sections describe the task formulation and each system component.

\subsection{Problem Formulation}

PresentAgent-2 addresses the task of query-to-presentation video generation. Given a natural-language user query $q$ and a presentation mode $m$, the system generates a presentation-style video $V_m$. The mode $m$ specifies one of three delivery forms: \emph{Single Presentation}, \emph{Discussion}, and \emph{Interaction}. Unlike document-to-presentation systems that start from a complete paper or report, our setting starts from a short and open-ended query, which usually lacks the full explanation content and visual resources needed for a presentation.

To obtain the missing context, PresentAgent-2 first summarizes the query into a focused topic $t$ and retrieves a multimodal resource set $\mathcal{R}$ through deep research:
\[
q \rightarrow (t, \mathcal{R}).
\]
Here, $\mathcal{R}$ denotes the retrieved multimodal resources, including text, images, GIFs, and videos. The system then generates the final presentation video based on the query, topic, retrieved resources, and selected mode:
\[
(q, t, \mathcal{R}, m) \rightarrow V_m.
\]
The presentation mode $m$ mainly determines the delivery script: single presentation uses single-speaker narration, discussion uses multi-speaker dialogue, and interaction uses an interactive presentation format.
\subsection{Deep Research for Multimodal Media}

To address the lack of content and visual materials in user queries, PresentAgent-2 uses deep research to collect textual information and multimodal media for the given query. Unlike standard search methods that mainly return plain text, our search is biased toward presentation-friendly sources, such as web pages, tutorials, demo pages, and articles with rich media or clear visual explanations.

Specifically, deep research first searches for a set of candidate URLs based on the extracted topic. The system then filters these URLs and prioritizes pages that are more suitable for constructing presentation content. The filtering criteria include two main aspects. First, the page should contain sufficiently complete textual content, rather than only short descriptions, title lists, or fragmented information. Second, the page should contain rich media resources, such as images, GIFs, or videos, to support a more intuitive and engaging presentation.

For the filtered URLs, the system further extracts the textual content and media resources to form a multimodal resource set. These materials are then used as input to the presentation generation stage, where the system determines which resources are suitable for the final slides and video.

\subsection{Presentation Generation}

In the presentation generation stage, PresentAgent-2 organizes the retrieved textual content and media resources into a set of presentation slides. The system first plans the presentation structure, including the overall outline, the topic of each slide, and how different resources should be used in the slides. Textual resources are used to generate slide titles, bullet points, and explanatory content, while image resources can be directly inserted into slides to support concept explanation, example illustration, and visual summarization.

For dynamic media such as GIFs and videos, PresentAgent-2 avoids simply converting them into static screenshots. Instead, during video composition, the system overlays the dynamic media onto the corresponding slide regions so that they remain playable in the final presentation video. In this way, the system can present dynamic processes, operation demonstrations, and visual examples within PPT-style pages, making the generated presentation video more vivid and engaging.

Meanwhile, the system generates a corresponding script for each slide and converts the script into audio. Finally, PresentAgent-2 composes the slide visuals, narration audio, and dynamic media into a complete presentation video. This allows the generated result to preserve the structured form of a presentation while using multimodal media to enhance visual communication.

\subsection{Three Presentation Modes}

PresentAgent-2 supports three presentation modes. These modes share the same deep research, presentation generation, and video composition pipeline, but differ in the script and delivery style.

\paragraph{Single Presentation.}
This mode generates a standard single-speaker presentation video. The system generates a narration script for each slide and delivers the content following the slide order.

\paragraph{Discussion.}
This mode turns the presentation content into a multi-speaker dialogue. Rather than simply splitting a single-speaker script into multiple parts, the system assigns different roles to the speakers, such as asking guiding questions, explaining concepts, clarifying details, and summarizing key points. This makes the presentation more conversational while keeping it grounded in the same slides and media.

\paragraph{Interaction.}
This mode extends the presentation with interactive question answering. The system can provide detailed answers to user questions, allowing the audience to participate during the presentation. The answers are grounded in the slides, scripts, and resources obtained through deep research, and the system can jump to the relevant slide when answering a question.

Additional implementation details are provided in Appendix~\ref{sec:app_implementation}.

\section{Experiments}
We evaluate PresentAgent-2 under the query-to-presentation video generation setting, focusing on its ability to support three presentation modes: Single Presentation, Discussion Presentation, and Interaction Presentation. Our experiments first compare PresentAgent-2 with representative related systems in terms of input settings, supported presentation modes, and multimodal resource support. We then evaluate the generated videos on our benchmark using objective quiz evaluation and subjective mode-specific evaluation, measuring both knowledge delivery and presentation quality.

\subsection{Evaluation Setting}

For automatic evaluation, we follow the protocol described in Section~\ref{sec:benchmark}. Each generated video is evaluated from two perspectives: objective knowledge delivery and subjective mode-specific quality. In objective quiz evaluation, the VLM acts as an audience member and answers five multiple-choice questions by watching the generated video and using the transcript transcribed from the generated video's audio, resulting in a quiz score from 0 to 5. In subjective evaluation, the VLM judge assigns independent 1--5 scores to each generated result according to the three metrics defined for the corresponding presentation mode. We report average quiz scores, the mean subjective score computed from the three mode-specific metrics, and the individual metric scores over examples for each mode and model.

\begin{table}[t]
\centering
\caption{
Capability comparison between PresentAgent-2 and representative related systems.
\checkmark indicates explicit support, $\triangle$ indicates partial or indirect support, and $\times$ indicates that the capability is not supported or not the target of the method.}
\label{tab:capability_comparison}
\resizebox{\linewidth}{!}{
\begin{tabular}{lccccccc}
\toprule
\textbf{Method} & \textbf{Presentation} & \textbf{Discussion} & \textbf{Interaction} & \textbf{Text} & \textbf{Image} & \textbf{GIF} & \textbf{Video} \\
\midrule
Paper2Video~\cite{zhu2025paper2video}  & \checkmark & $\times$ & $\times$ & \checkmark & \checkmark & $\times$ & $\triangle$ \\
Paper2Poster~\cite{pang2025paper2poster} & $\triangle$ & $\times$ & $\times$ & \checkmark & \checkmark & $\times$ & $\times$ \\
VideoDirectorGPT~\cite{lin2023videodirectorgpt}           & $\times$ & $\times$ & $\times$ & $\triangle$ & $\times$ & $\times$ & $\times$ \\
VideoStudio~\cite{long2024videostudio} & $\times$ & $\times$ & $\times$ & $\triangle$ & $\times$ & $\times$ & $\times$ \\
LVD~\cite{lian2023llm}                         & $\times$ & $\times$ & $\times$ & $\triangle$ & $\times$ & $\times$ & $\times$ \\
PresentAgent~\cite{shi2025presentagent}             & \checkmark & $\times$ & $\times$ & \checkmark & $\triangle$ & $\times$ & $\times$ \\
\midrule
\textbf{PresentAgent-2}     & \checkmark & \checkmark &  \checkmark  & \checkmark & \checkmark & \checkmark & \checkmark \\
\bottomrule
\end{tabular}
}
\end{table}

\subsection{Main Results}

\paragraph{Capability Analysis.}
Table~\ref{tab:capability_comparison} compares PresentAgent-2 with representative related systems in terms of task setting and supported capabilities. Existing systems are typically designed for more limited task settings, such as document-to-presentation, paper-to-video, poster generation, or general video generation. In contrast, PresentAgent-2 targets the more open-ended query-to-presentation setting, where the system starts from a user query, performs deep research, and generates presentation videos across different delivery modes. It supports Single Presentation, Discussion Presentation, and Interaction Presentation, while also integrating text, images, GIFs, and video clips as embedded media resources. This capability coverage shows that PresentAgent-2 moves beyond single-format presentation generation toward a more general multimodal presentation agent.

\paragraph{Benchmark Evaluation.}
Table~\ref{tab:benchmark_results} reports the benchmark evaluation results of PresentAgent-2. 
With the Qwen3.5-VL-Plus backbone, PresentAgent-2 achieves quiz scores of 4.84, 4.85, and 4.85 on Single Presentation, Discussion Presentation, and Interaction Presentation, respectively. 
It also obtains mean subjective scores of 4.47, 4.37, and 4.52 across the three modes, showing that PresentAgent-2 can convey key knowledge and generate mode-aware presentation videos from user queries.
\begin{table*}[t]
  \caption{
    Benchmark evaluation results of Human Reference and PresentAgent-2 with different models.
    Quiz is averaged on a 0--5 scale, and subjective scores are on a 1--5 scale.
    Metric abbreviations follow Table~\ref{tab:subjective_metrics}.
    }
  \label{tab:benchmark_results}
  \centering
  \scriptsize
  \setlength{\tabcolsep}{3pt}
  \renewcommand{\arraystretch}{1.08}
  \resizebox{\textwidth}{!}{
  \begin{tabular}{cc|ccccc|ccccc|ccccc}
    \toprule
    \multirow{2}{*}{\textbf{Method}}
    & \multirow{2}{*}{\textbf{Model}}
    & \multicolumn{5}{c|}{\textbf{Single Presentation Score}}
    & \multicolumn{5}{c|}{\textbf{Discussion Presentation Score}}
    & \multicolumn{5}{c}{\textbf{Interaction Presentation Score}} \\
    \cmidrule(lr){3-7}
    \cmidrule(lr){8-12}
    \cmidrule(lr){13-17}
    &
    & \textbf{Quiz}
    & \textbf{QA}
    & \textbf{DRE}
    & \textbf{VDQ}
    & \textbf{Mean}
    & \textbf{Quiz}
    & \textbf{DE}
    & \textbf{SRC}
    & \textbf{CD}
    & \textbf{Mean}
    & \textbf{Quiz}
    & \textbf{AE}
    & \textbf{CC}
    & \textbf{IH}
    & \textbf{Mean} \\
    \midrule

    Human Reference
    & Human-created
    & 4.82 & 4.54 & 4.45 & 4.40 & 4.46
    & 4.83 & 4.45 & 4.31 & 4.45 & 4.40
    & -- & -- & -- & -- & -- \\

    \midrule

    PresentAgent-2
    & Qwen3.5-VL-Plus
    & 4.84 & 4.50 & 4.48 & 4.43 & 4.47
    & 4.85 & 4.43 & 4.22 & 4.47 & 4.37
    & 4.85 & 4.65 & 4.43 & 4.49 & 4.52 \\

    PresentAgent-2
    & Claude Opus 4.7
    & 4.80 & 4.47 & 4.47 & 4.35 & 4.43
    & 4.82 & 4.43 & 4.21 & 4.49 & 4.38
    & 4.80 & 4.55 & 4.55 & 4.47 & 4.52 \\

    PresentAgent-2
    & Gemini 3.1 Pro
    & 4.78 & 4.45 & 4.24 & 4.37 & 4.35
    & 4.80 & 4.37 & 4.08 & 4.30 & 4.25
    & 4.75 & 4.53 & 4.40 & 4.42 & 4.45 \\

    PresentAgent-2
    & GPT-5.5
    & 4.83 & 4.25 & 4.30 & 4.20 & 4.25
    & 4.77 & 4.19 & 4.11 & 4.20 & 4.17
    & 4.75 & 4.54 & 4.35 & 4.50 & 4.46 \\

    PresentAgent-2
    & GLM-4.7V
    & 4.75 & 4.21 & 4.21 & 4.13 & 4.18
    & 4.67 & 4.20 & 4.03 & 4.11 & 4.11
    & 4.60 & 4.53 & 4.33 & 4.41 & 4.42 \\

    \bottomrule
  \end{tabular}
  }
\end{table*}

The mode-specific subjective metrics further reveal how PresentAgent-2 adapts to different presentation settings. For Single Presentation, the system organizes retrieved textual and multimodal resources into coherent explanatory videos. For Discussion Presentation, it reformulates technical content into multi-speaker dialogue with complementary speaker roles and natural conversational delivery. For Interaction Presentation, the generated context supports effective, comprehensible, and helpful answers to audience questions.
\begin{figure*}[t]
\centering
\scriptsize
\setlength{\tabcolsep}{2pt}
\renewcommand{\arraystretch}{1.0}

\setlength{\fboxsep}{0pt}
\setlength{\fboxrule}{0.35pt}

\newcommand{\figwidth}{0.238\textwidth}
\newcommand{\rowgap}{0.75em}

\newcommand{\imgcell}[1]{%
\begin{minipage}[t]{\figwidth}
    \centering
    \fbox{%
        \includegraphics[
            width=\dimexpr\linewidth-2\fboxsep-2\fboxrule\relax
        ]{#1}%
    }
\end{minipage}
}

\newcommand{\rowtitle}[1]{%
\multicolumn{4}{c}{\footnotesize\textbf{#1}}\\[-0.15em]
}

\begin{tabular}{@{}cccc@{}}

\rowtitle{Single Presentation}
\imgcell{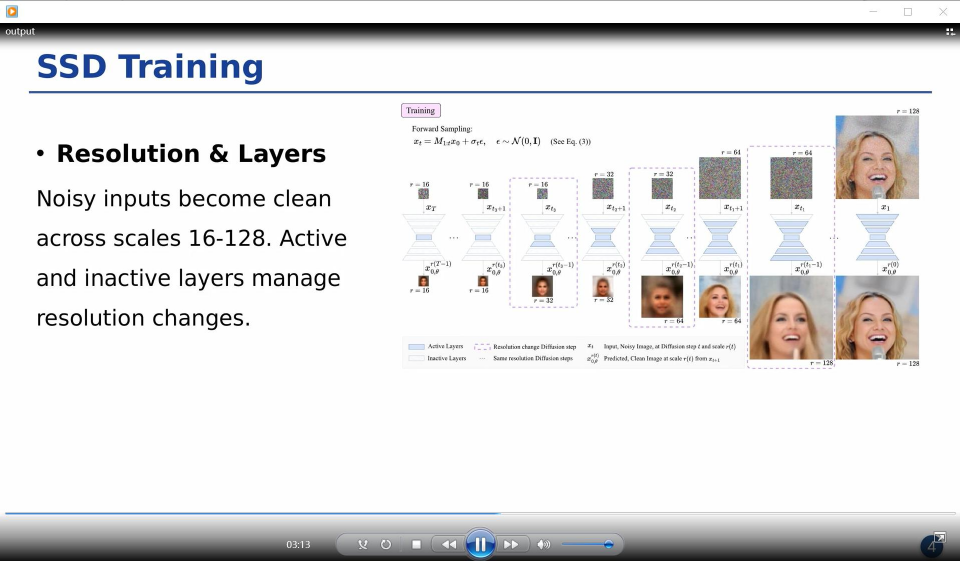}
&
\imgcell{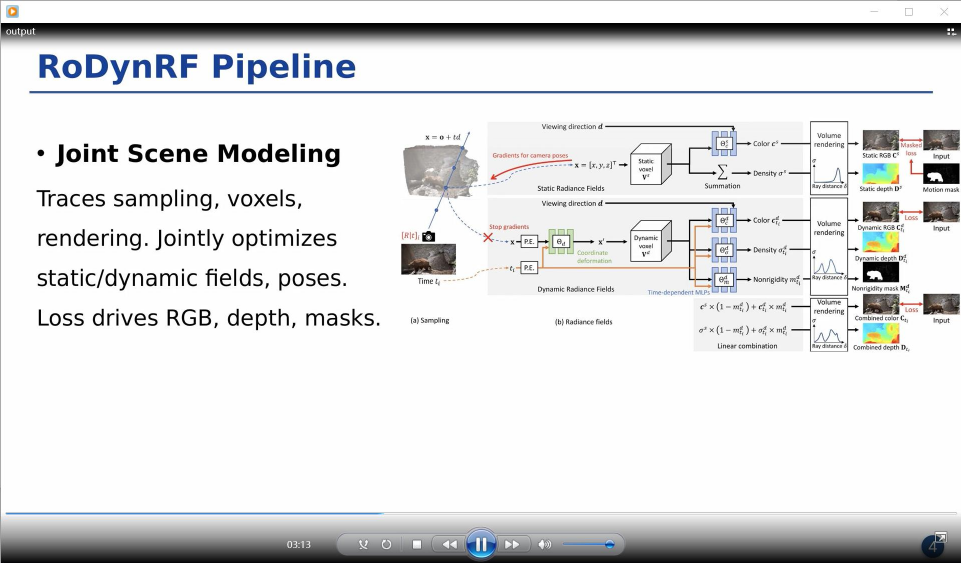}
&
\imgcell{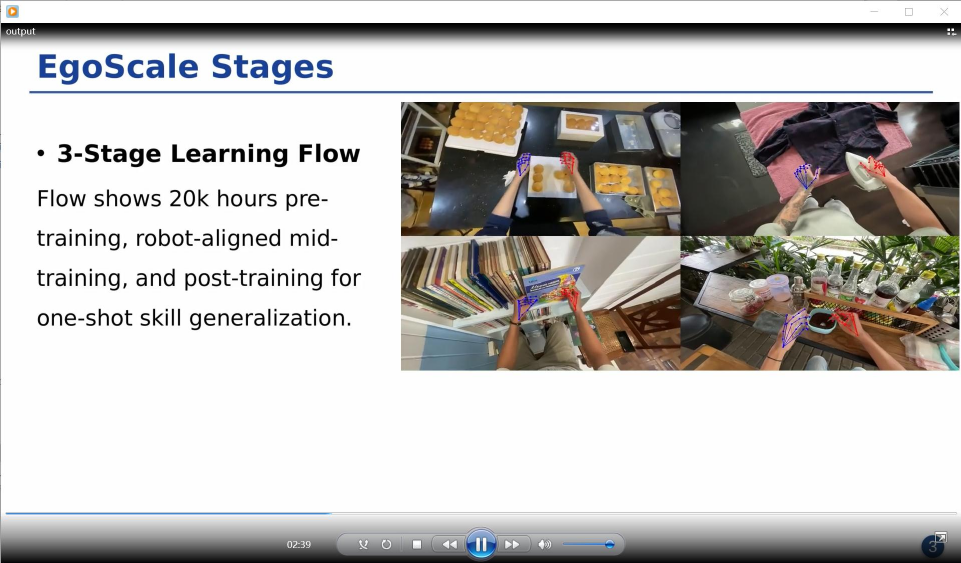}
&
\imgcell{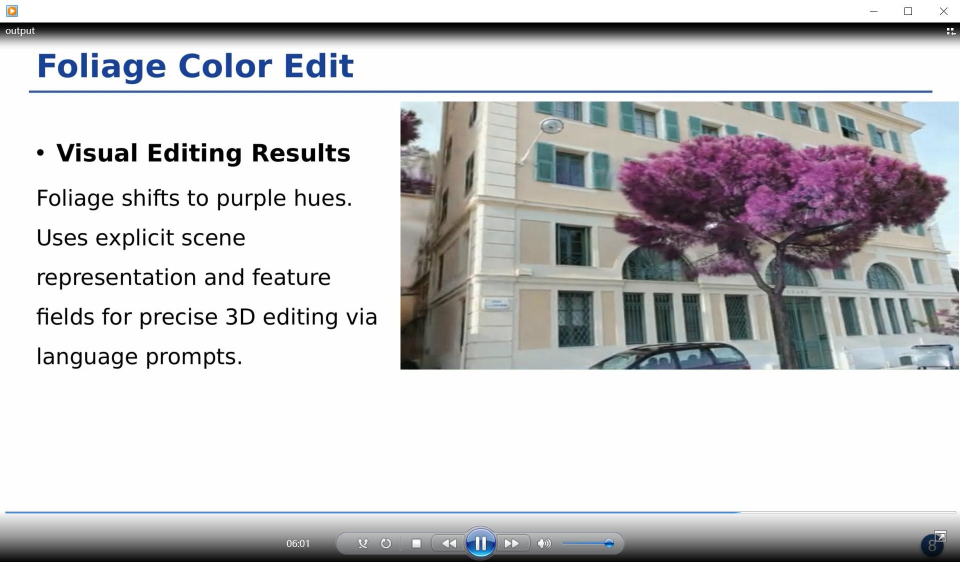}
\\
\noalign{\vskip \rowgap}

\rowtitle{Discussion Presentation}
\imgcell{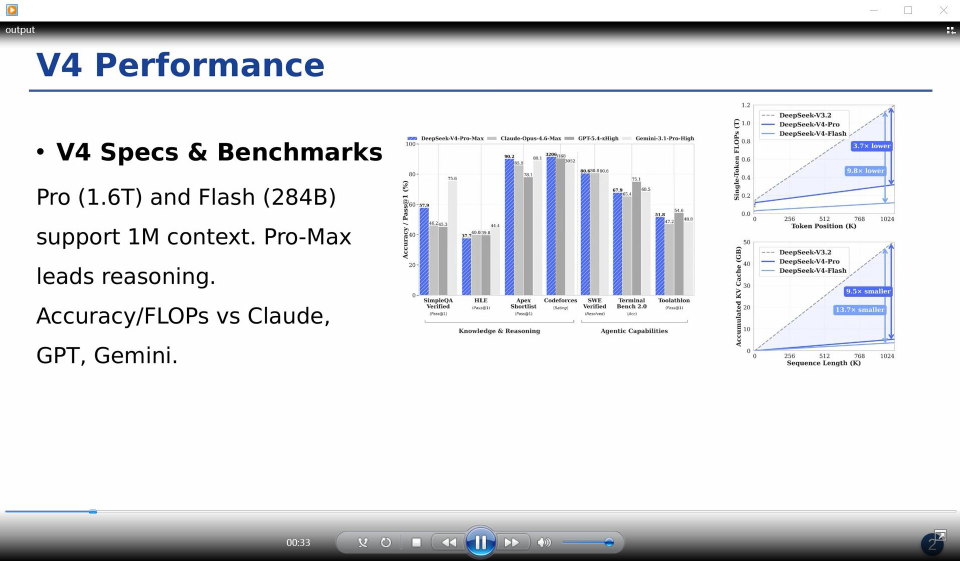}
&
\imgcell{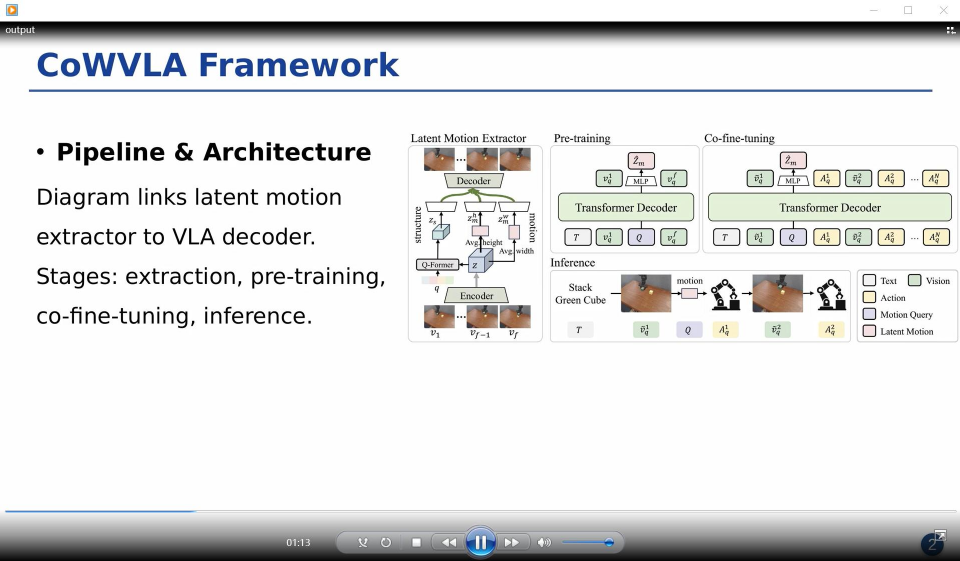}
&
\imgcell{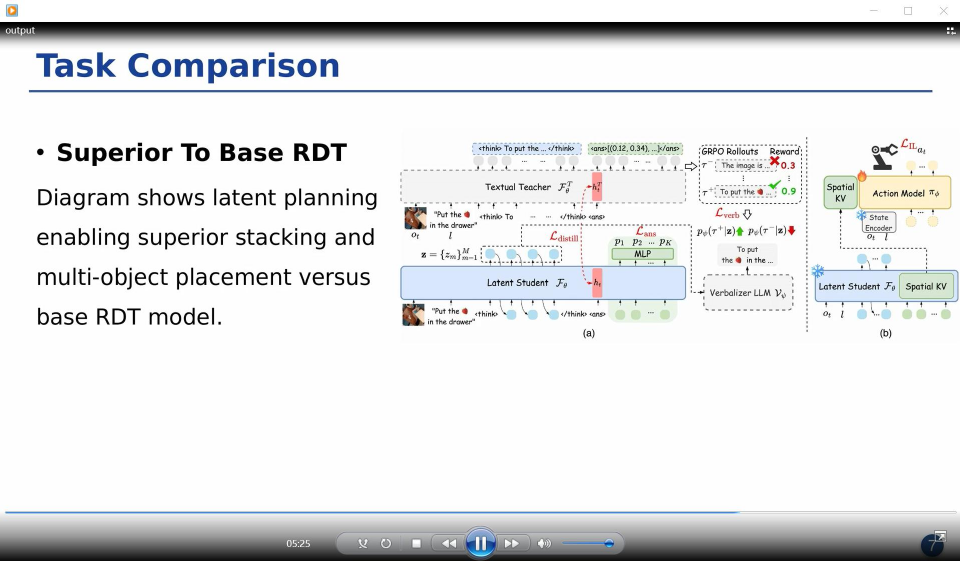}
&
\imgcell{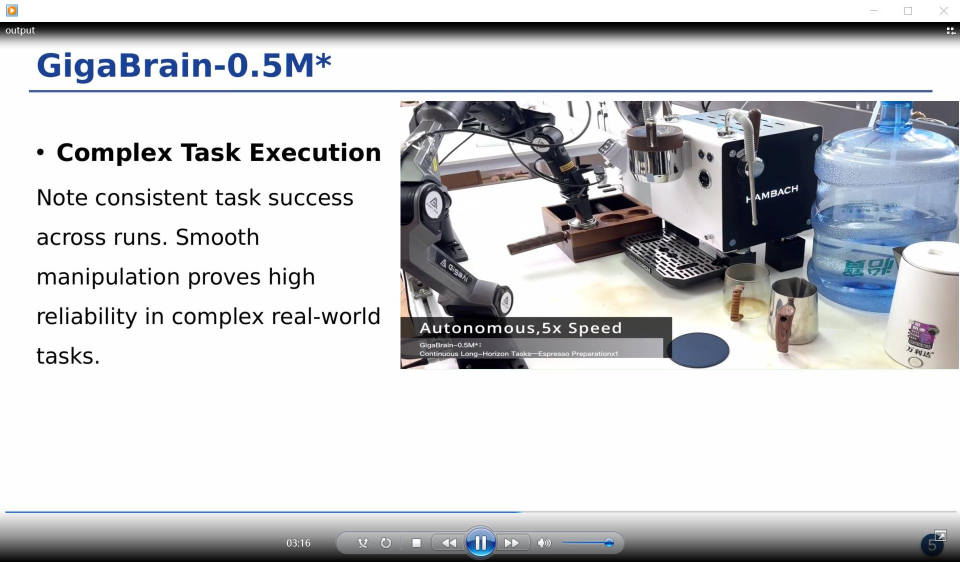}
\\
\noalign{\vskip \rowgap}

\rowtitle{Interaction Presentation}
\imgcell{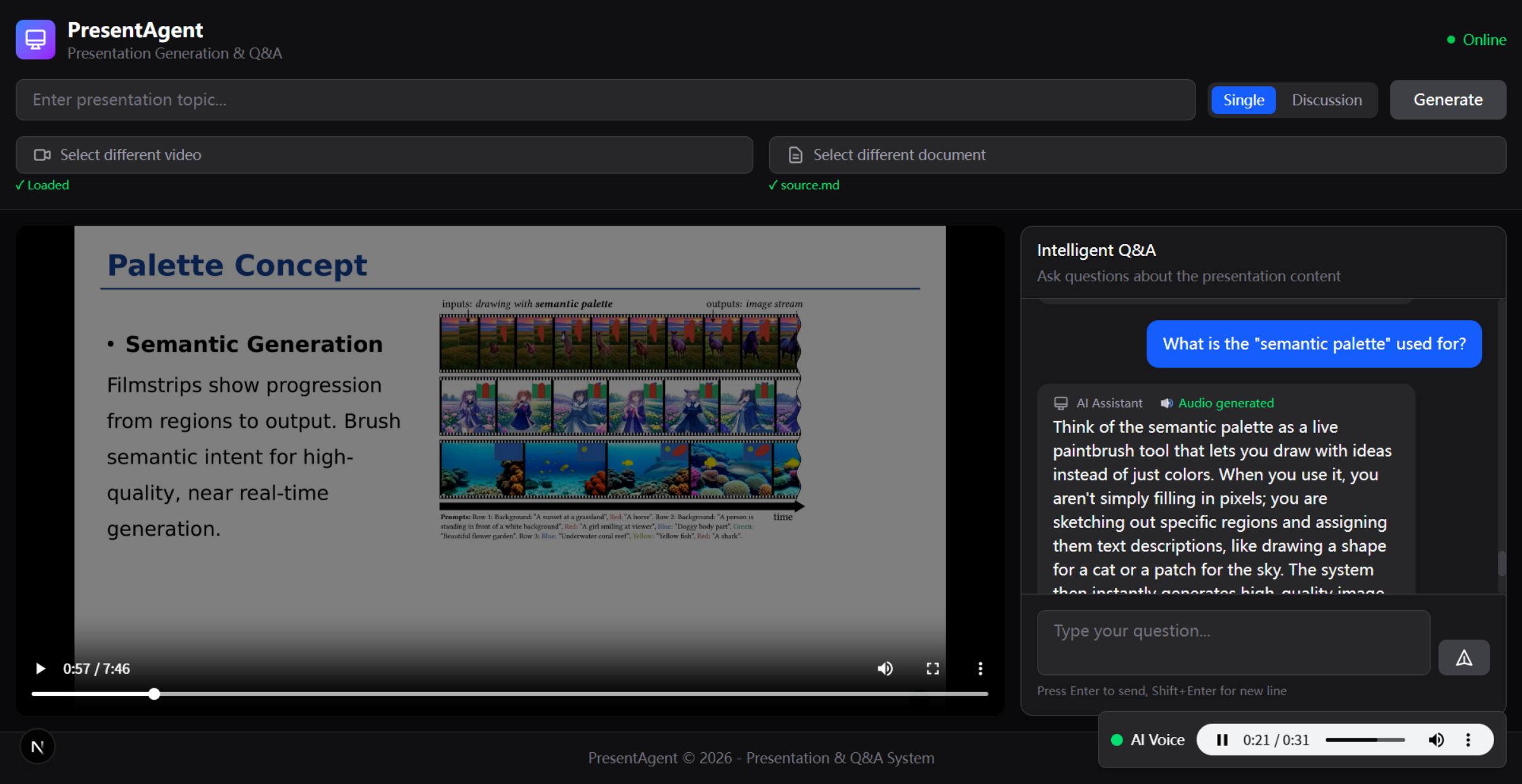}
&
\imgcell{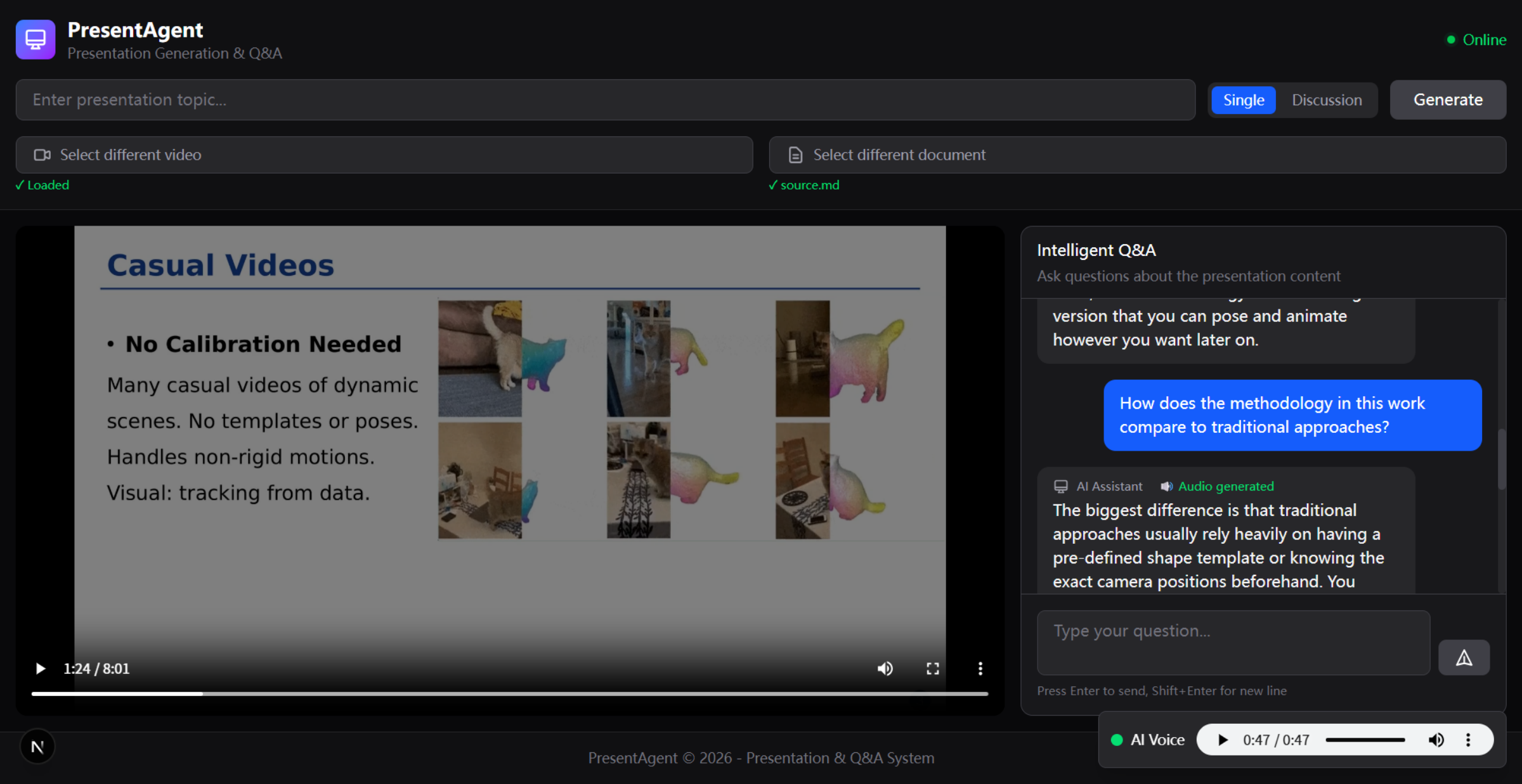}
&
\imgcell{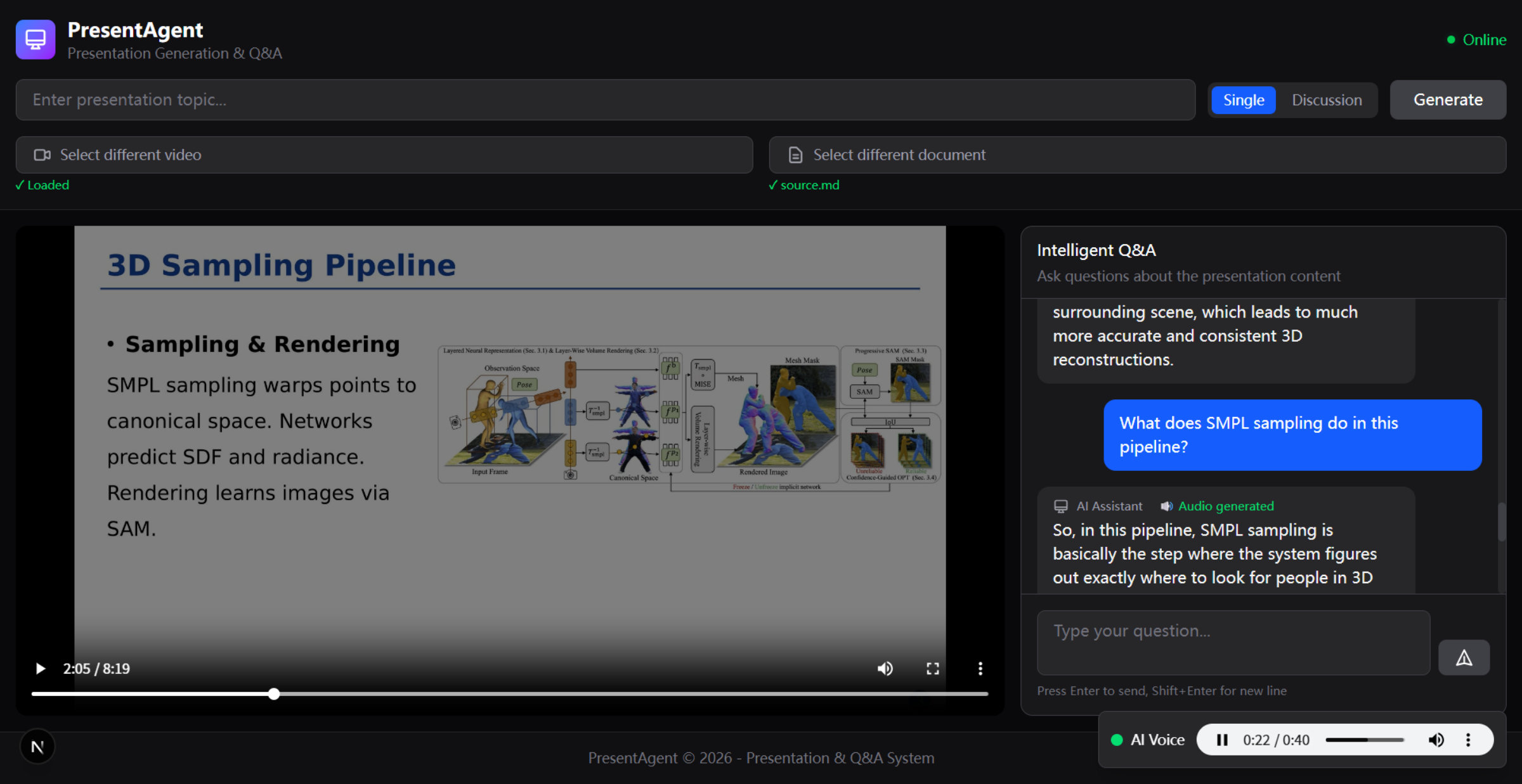}
&
\imgcell{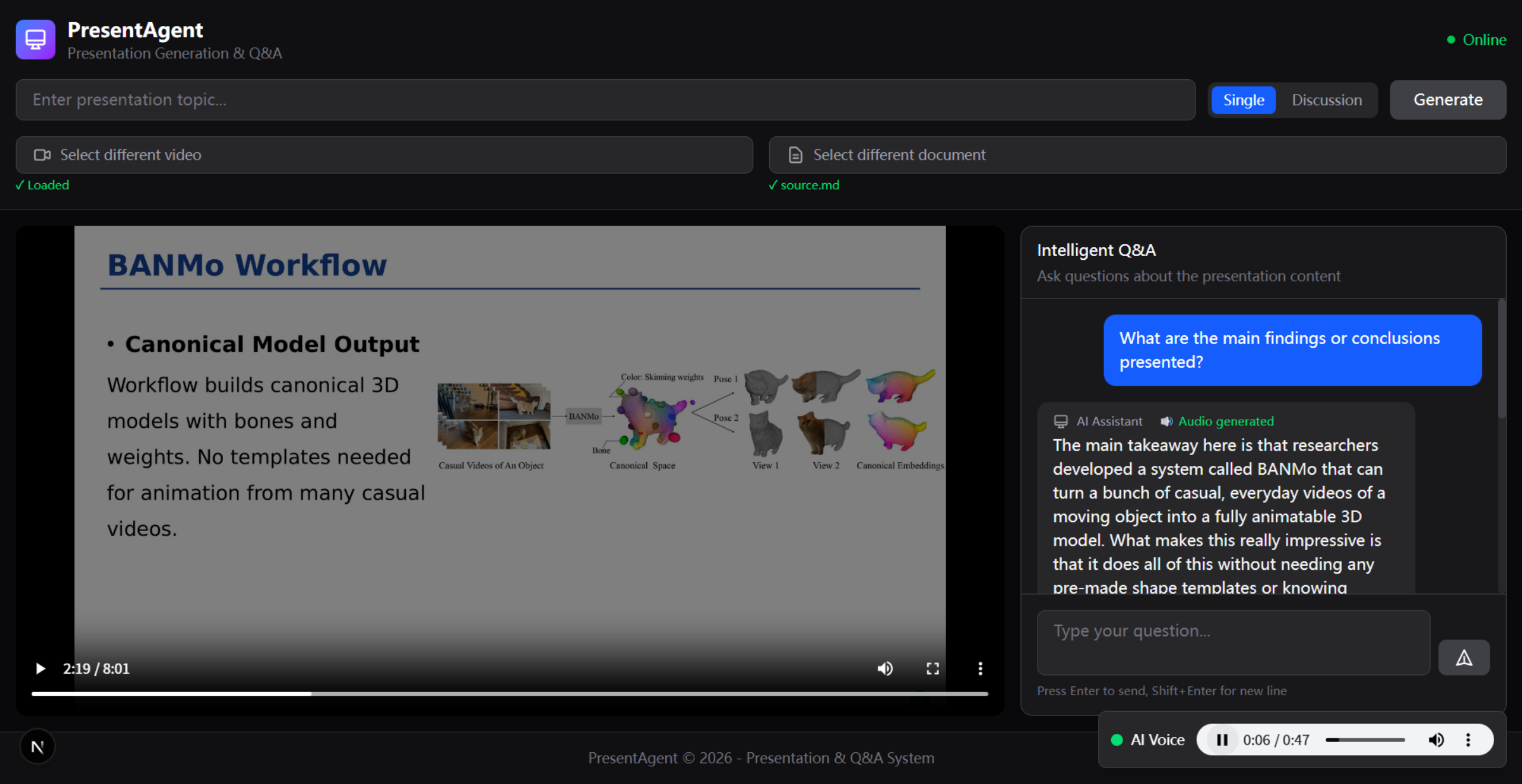}
\\

\end{tabular}

\caption{
Qualitative examples of PresentAgent-2 across three presentation settings.
Rows from top to bottom show Single Presentation, Discussion Presentation, and Interaction Presentation, respectively.
All panels are video frames generated by PresentAgent-2.
}
\label{fig:qualitative_examples}
\end{figure*}
\paragraph{Qualitative Demonstrations.}
Figure~\ref{fig:qualitative_examples} shows representative generated examples across different presentation modes. For Single Presentation, PresentAgent-2 produces a structured explanation that combines slides, narration, and retrieved visual evidence. For Discussion Presentation, the system reformulates similar technical content into a multi-speaker dialogue, in which speakers ask questions, compare concepts, and summarize key points. For Interaction Presentation, the generated interface supports audience questions after the presentation, and the system answers them based on the generated presentation context. These qualitative examples make the mode-specific behaviors visible and complement the quantitative results in Table~\ref{tab:benchmark_results}. We provide additional key-frame visualizations, generated slide/script examples, and Interaction Presentation screenshots in Appendix~\ref{sec:appendix_qualitative_examples}.

\subsection{Analysis}

Figure~\ref{fig:qualitative_examples} shows representative outputs of PresentAgent-2 across different presentation modes. Single Presentation provides structured explanatory videos, Discussion Presentation reformulates content into multi-speaker dialogue, and Interaction Presentation supports audience-facing question answering based on the generated presentation context. Together with the benchmark results in Table~\ref{tab:benchmark_results}, these examples show that PresentAgent-2 can generate coherent, informative, and mode-aware presentation videos from open-ended user queries. Additional ablation studies in Appendix~\ref{app:ablation} analyze the effects of multimodal resource usage, dynamic media preservation, and role-aware discussion generation.

\section{Conclusion}
We present PresentAgent-2, a query-to-presentation video generation agent that transforms open-ended user queries into multimodal presentation videos. The framework integrates deep research, multimodal resource retrieval, slide/script generation, audio synthesis, and video composition, and supports Single Presentation, Discussion Presentation, and Interaction Presentation. We introduce PresentEval to evaluate objective knowledge delivery and subjective mode-specific quality. Experiments show that PresentAgent-2 generates informative and mode-aware presentation videos. Limitations are discussed in Appendix~\ref{sec:app_limitations}.

\clearpage
\begin{ack}
This research was undertaken with the assistance of resources from the National Computational Infrastructure (NCI Australia), an NCRIS enabled capability supported by the Australian Government. Furthermore, this work was supported by the Fundamental Research Funds for the Central Universities, Peking University.
\end{ack}

\bibliographystyle{unsrt}
\bibliography{paper/refs}

@inproceedings{shi2025presentagent,
  title={Presentagent: Multimodal agent for presentation video generation},
  author={Shi, Jingwei and Zhang, Zeyu and Wu, Biao and Liang, Yanjie and Fang, Meng and Chen, Ling and Zhao, Yang},
  booktitle={Proceedings of the 2025 Conference on Empirical Methods in Natural Language Processing: System Demonstrations},
  pages={760--773},
  year={2025}
}

@article{zhu2025paper2video,
  title={Paper2video: Automatic video generation from scientific papers},
  author={Zhu, Zeyu and Lin, Kevin Qinghong and Shou, Mike Zheng},
  journal={arXiv preprint arXiv:2510.05096},
  year={2025}
}

@article{yang2025auto,
  title={Auto-slides: An interactive multi-agent system for creating and customizing research presentations},
  author={Yang, Yuheng and Jiang, Wenjia and Wang, Yang and Song, Yi and Wang, Yiwei and Zhang, Chi},
  journal={arXiv preprint arXiv:2509.11062},
  year={2025}
}

@article{jung2025talk,
  title={Talk to your slides: Language-driven agents for efficient slide editing},
  author={Jung, Kyudan and Cho, Hojun and Yun, Jooyeol and Yang, Soyoung and Jang, Jaehyeok and Choo, Jaegul},
  journal={arXiv preprint arXiv:2505.11604},
  year={2025}
}

@inproceedings{zheng2025pptagent,
  title={Pptagent: Generating and evaluating presentations beyond text-to-slides},
  author={Zheng, Hao and Guan, Xinyan and Kong, Hao and Zhang, Wenkai and Zheng, Jia and Zhou, Weixiang and Lin, Hongyu and Lu, Yaojie and Han, Xianpei and Sun, Le},
  booktitle={Proceedings of the 2025 Conference on Empirical Methods in Natural Language Processing},
  pages={14413--14429},
  year={2025}
}

@article{xu2025pregenie,
  title={PreGenie: An Agentic Framework for High-quality Visual Presentation Generation},
  author={Xu, Xiaojie and Xu, Xinli and Chen, Sirui and Chen, Haoyu and Zhang, Fan and Chen, Ying-Cong},
  journal={arXiv preprint arXiv:2505.21660},
  year={2025}
}

@article{liang2025slidegen,
  title={Slidegen: Collaborative multimodal agents for scientific slide generation},
  author={Liang, Xin and Zhang, Xiang and Xu, Yiwei and Sun, Siqi and You, Chenyu},
  journal={arXiv preprint arXiv:2512.04529},
  year={2025}
}

@article{chen2025presentcoach,
  title={PresentCoach: Dual-Agent Presentation Coaching through Exemplars and Interactive Feedback},
  author={Chen, Sirui and Zhou, Jinsong and Xu, Xinli and Yang, Xiaoyu and Guo, Litao and Chen, Ying-Cong},
  journal={arXiv preprint arXiv:2511.15253},
  year={2025}
}

@article{liu2025presenting,
  title={Presenting a Paper is an Art: Self-Improvement Aesthetic Agents for Academic Presentations},
  author={Liu, Chengzhi and Yang, Yuzhe and Zhou, Kaiwen and Zhang, Zhen and Fan, Yue and Xie, Yanan and Qi, Peng and Wang, Xin Eric},
  journal={arXiv preprint arXiv:2510.05571},
  year={2025}
}

@article{liang2025videoagent,
  title={VideoAgent: Personalized Synthesis of Scientific Videos},
  author={Liang, Xiao and Li, Bangxin and Chen, Zixuan and Zheng, Hanyue and Ma, Zhi and Wang, Di and Tian, Cong and Wang, Quan},
  journal={arXiv preprint arXiv:2509.11253},
  year={2025}
}

@article{kong2025let,
  title={Let them talk: Audio-driven multi-person conversational video generation},
  author={Kong, Zhe and Gao, Feng and Zhang, Yong and Kang, Zhuoliang and Wei, Xiaoming and Cai, Xunliang and Chen, Guanying and Luo, Wenhan},
  journal={arXiv preprint arXiv:2505.22647},
  year={2025}
}

@article{hu2025polyvivid,
  title={PolyVivid: Vivid Multi-Subject Video Generation with Cross-Modal Interaction and Enhancement},
  author={Hu, Teng and Yu, Zhentao and Zhou, Zhengguang and Zhang, Jiangning and Zhou, Yuan and Lu, Qinglin and Yi, Ran},
  journal={arXiv preprint arXiv:2506.07848},
  year={2025}
}

@article{kyaw2025node,
  title={Node-Based Editing for Multimodal Generation of Text, Audio, Image, and Video},
  author={Kyaw, Alexander Htet and Sivalingam, Lenin Ravindranath},
  journal={arXiv preprint arXiv:2511.03227},
  year={2025}
}

@article{pang2025paper2poster,
  title={Paper2poster: Towards multimodal poster automation from scientific papers},
  author={Pang, Wei and Lin, Kevin Qinghong and Jian, Xiangru and He, Xi and Torr, Philip},
  journal={arXiv preprint arXiv:2505.21497},
  year={2025}
}

@inproceedings{fu2022doc2ppt,
  title={Doc2ppt: Automatic presentation slides generation from scientific documents},
  author={Fu, Tsu-Jui and Wang, William Yang and McDuff, Daniel and Song, Yale},
  booktitle={Proceedings of the AAAI Conference on Artificial Intelligence},
  volume={36},
  number={1},
  pages={634--642},
  year={2022}
}

@inproceedings{sun2025genesis,
  title={Os-genesis: Automating gui agent trajectory construction via reverse task synthesis},
  author={Sun, Qiushi and Cheng, Kanzhi and Ding, Zichen and Jin, Chuanyang and Wang, Yian and Xu, Fangzhi and Wu, Zhenyu and Jia, Chengyou and Chen, Liheng and Liu, Zhoumianze and others},
  booktitle={Proceedings of the 63rd Annual Meeting of the Association for Computational Linguistics (Volume 1: Long Papers)},
  pages={5555--5579},
  year={2025}
}

@inproceedings{wang2026mavis,
  title={MAViS: A multi-agent framework for long-sequence video storytelling},
  author={Wang, Qian and Huang, Ziqi and Jia, Ruoxi and Debevec, Paul and Yu, Ning},
  booktitle={Proceedings of the 19th Conference of the European Chapter of the Association for Computational Linguistics (Volume 1: Long Papers)},
  pages={2273--2295},
  year={2026}
}

@article{yang2023gpt4tools,
  title={Gpt4tools: Teaching large language model to use tools via self-instruction},
  author={Yang, Rui and Song, Lin and Li, Yanwei and Zhao, Sijie and Ge, Yixiao and Li, Xiu and Shan, Ying},
  journal={Advances in Neural Information Processing Systems},
  volume={36},
  pages={71995--72007},
  year={2023}
}

@article{yang2023mm,
  title={Mm-react: Prompting chatgpt for multimodal reasoning and action},
  author={Yang, Zhengyuan and Li, Linjie and Wang, Jianfeng and Lin, Kevin and Azarnasab, Ehsan and Ahmed, Faisal and Liu, Zicheng and Liu, Ce and Zeng, Michael and Wang, Lijuan},
  journal={arXiv preprint arXiv:2303.11381},
  year={2023}
}

@article{zhang2025motion,
  title={Motion anything: Any to motion generation},
  author={Zhang, Zeyu and Wang, Yiran and Mao, Wei and Li, Danning and Zhao, Rui and Wu, Biao and Song, Zirui and Zhuang, Bohan and Reid, Ian and Hartley, Richard},
  journal={arXiv preprint arXiv:2503.06955},
  year={2025}
}

@article{zhang2024infinimotion,
  title={Infinimotion: Mamba boosts memory in transformer for arbitrary long motion generation},
  author={Zhang, Zeyu and Liu, Akide and Chen, Qi and Chen, Feng and Reid, Ian and Hartley, Richard and Zhuang, Bohan and Tang, Hao},
  journal={arXiv preprint arXiv:2407.10061},
  year={2024}
}

@article{zhang2024kmm,
  title={Kmm: Key frame mask mamba for extended motion generation},
  author={Zhang, Zeyu and Gao, Hang and Liu, Akide and Chen, Qi and Chen, Feng and Wang, Yiran and Li, Danning and Zhao, Rui and Li, Zhenming and Zhou, Zhongwen and others},
  journal={arXiv preprint arXiv:2411.06481},
  year={2024}
}

@inproceedings{li2023evaluating,
  title={Evaluating object hallucination in large vision-language models},
  author={Li, Yifan and Du, Yifan and Zhou, Kun and Wang, Jinpeng and Zhao, Xin and Wen, Ji-Rong},
  booktitle={Proceedings of the 2023 conference on empirical methods in natural language processing},
  pages={292--305},
  year={2023}
}

@article{wu2025towards,
  title={Towards generalist foundation model for radiology by leveraging web-scale 2d\&3d medical data},
  author={Wu, Chaoyi and Zhang, Xiaoman and Zhang, Ya and Hui, Hui and Wang, Yanfeng and Xie, Weidi},
  journal={Nature Communications},
  volume={16},
  number={1},
  pages={7866},
  year={2025},
  publisher={Nature Publishing Group UK London}
}

@article{team2026qwen3,
  title={Qwen3. 5-omni technical report},
  author={Team, Qwen},
  journal={arXiv preprint arXiv:2604.15804},
  year={2026}
}

@inproceedings{ge2025autopresent,
  title={Autopresent: Designing structured visuals from scratch},
  author={Ge, Jiaxin and Wang, Zora Zhiruo and Zhou, Xuhui and Peng, Yi-Hao and Subramanian, Sanjay and Tan, Qinyue and Sap, Maarten and Suhr, Alane and Fried, Daniel and Neubig, Graham and others},
  booktitle={Proceedings of the Computer Vision and Pattern Recognition Conference},
  pages={2902--2911},
  year={2025}
}

@article{wang2025infinity,
  title={Infinity parser: Layout aware reinforcement learning for scanned document parsing},
  author={Wang, Baode and Wu, Biao and Li, Weizhen and Fang, Meng and Huang, Zuming and Huang, Jun and Wang, Haozhe and Liang, Yanjie and Chen, Ling and Chu, Wei and others},
  journal={arXiv preprint arXiv:2506.03197},
  year={2025}
}

@article{li2023videogen,
  title={Videogen: A reference-guided latent diffusion approach for high definition text-to-video generation},
  author={Li, Xin and Chu, Wenqing and Wu, Ye and Yuan, Weihang and Liu, Fanglong and Zhang, Qi and Li, Fu and Feng, Haocheng and Ding, Errui and Wang, Jingdong},
  journal={arXiv preprint arXiv:2309.00398},
  year={2023}
}

@inproceedings{xue2025phyt2v,
  title={Phyt2v: Llm-guided iterative self-refinement for physics-grounded text-to-video generation},
  author={Xue, Qiyao and Yin, Xiangyu and Yang, Boyuan and Gao, Wei},
  booktitle={Proceedings of the Computer Vision and Pattern Recognition Conference},
  pages={18826--18836},
  year={2025}
}

@article{yang2024cogvideox,
  title={Cogvideox: Text-to-video diffusion models with an expert transformer},
  author={Yang, Zhuoyi and Teng, Jiayan and Zheng, Wendi and Ding, Ming and Huang, Shiyu and Xu, Jiazheng and Yang, Yuanming and Hong, Wenyi and Zhang, Xiaohan and Feng, Guanyu and others},
  journal={arXiv preprint arXiv:2408.06072},
  year={2024}
}

@inproceedings{zhang2024motion,
  title={Motion mamba: Efficient and long sequence motion generation},
  author={Zhang, Zeyu and Liu, Akide and Reid, Ian and Hartley, Richard and Zhuang, Bohan and Tang, Hao},
  booktitle={European Conference on Computer Vision},
  pages={265--282},
  year={2024},
  organization={Springer}
}

@article{deng2025emerging,
  title={Emerging properties in unified multimodal pretraining},
  author={Deng, Chaorui and Zhu, Deyao and Li, Kunchang and Gou, Chenhui and Li, Feng and Wang, Zeyu and Zhong, Shu and Yu, Weihao and Nie, Xiaonan and Song, Ziang and others},
  journal={arXiv preprint arXiv:2505.14683},
  year={2025}
}

@article{xie2024show,
  title={Show-o: One single transformer to unify multimodal understanding and generation},
  author={Xie, Jinheng and Mao, Weijia and Bai, Zechen and Zhang, David Junhao and Wang, Weihao and Lin, Kevin Qinghong and Gu, Yuchao and Chen, Zhijie and Yang, Zhenheng and Shou, Mike Zheng},
  journal={arXiv preprint arXiv:2408.12528},
  year={2024}
}

@inproceedings{lin2025showui,
  title={Showui: One vision-language-action model for gui visual agent},
  author={Lin, Kevin Qinghong and Li, Linjie and Gao, Difei and Yang, Zhengyuan and Wu, Shiwei and Bai, Zechen and Lei, Stan Weixian and Wang, Lijuan and Shou, Mike Zheng},
  booktitle={Proceedings of the Computer Vision and Pattern Recognition Conference},
  pages={19498--19508},
  year={2025}
}

@inproceedings{hu2025multimodal,
  title={Multimodal Content Alignment with LLM for Visual Presentation of Papers},
  author={Hu, Huiying and He, Zhicheng and Zhou, Yixiao and Zhang, Tongwei and Lyu, Xiaoqing},
  booktitle={International Conference on Document Analysis and Recognition},
  pages={238--256},
  year={2025},
  organization={Springer}
}

@inproceedings{konstantinov2026slides,
  title={Slides Agent: An Intelligent Agent for Creating and Analyzing Presentations Using Large},
  author={Konstantinov, Aleksandr and Avdyushina, Anna and Markina, Tatiana},
  booktitle={Creativity in Intelligent Technologies and Data Science: 6th International Conference, CIT\&DS 2025, Volgograd, Russia, September 22--25, 2025, Proceedings},
  pages={123},
  year={2026},
  organization={Springer Nature}
}

@article{zhao2025unified,
  title={Unified multimodal understanding and generation models: Advances, challenges, and opportunities},
  author={Zhao, Shanshan and Zhang, Xinjie and Guo, Jintao and Hu, Jiakui and Duan, Lunhao and Fu, Minghao and Chng, Yong Xien and Wang, Guo-Hua and Chen, Qing-Guo and Xu, Zhao and others},
  journal={arXiv preprint arXiv:2505.02567},
  year={2025}
}

@article{lin2023videodirectorgpt,
  title={Videodirectorgpt: Consistent multi-scene video generation via llm-guided planning},
  author={Lin, Han and Zala, Abhay and Cho, Jaemin and Bansal, Mohit},
  journal={arXiv preprint arXiv:2309.15091},
  year={2023}
}

@inproceedings{long2024videostudio,
  title={Videostudio: Generating consistent-content and multi-scene videos},
  author={Long, Fuchen and Qiu, Zhaofan and Yao, Ting and Mei, Tao},
  booktitle={European Conference on Computer Vision},
  pages={468--485},
  year={2024},
  organization={Springer}
}

@article{lian2023llm,
  title={Llm-grounded video diffusion models},
  author={Lian, Long and Shi, Baifeng and Yala, Adam and Darrell, Trevor and Li, Boyi},
  journal={arXiv preprint arXiv:2309.17444},
  year={2023}
}

\clearpage
\appendix
\section{Additional Qualitative Examples}
\label{sec:appendix_qualitative_examples}

As shown in Figures~\ref{fig:appendix_examples_01} and~\ref{fig:appendix_examples_02}, we provide additional qualitative examples of PresentAgent-2.
Each column corresponds to one example.
Rows from top to bottom show a representative Single Presentation frame, the generated slide, the generated script, and a representative Interaction Presentation screenshot.

\newcommand{\appcell}[1]{%
\fbox{%
\begin{minipage}[c][2.35cm][c]{0.315\textwidth}
\centering
\includegraphics[
    width=\linewidth,
    height=2.25cm,
    keepaspectratio
]{#1}%
\end{minipage}%
}%
}

\newcommand{\appRowGap}{0.65em}

\begin{figure*}[h]
\centering
\scriptsize
\setlength{\tabcolsep}{3pt}
\renewcommand{\arraystretch}{1.0}
\setlength{\fboxsep}{0pt}
\setlength{\fboxrule}{0.35pt}

\begin{tabular}{@{}ccc@{}}
\textbf{Example 1}
&
\textbf{Example 2}
&
\textbf{Example 3}
\\[0.25em]

\appcell{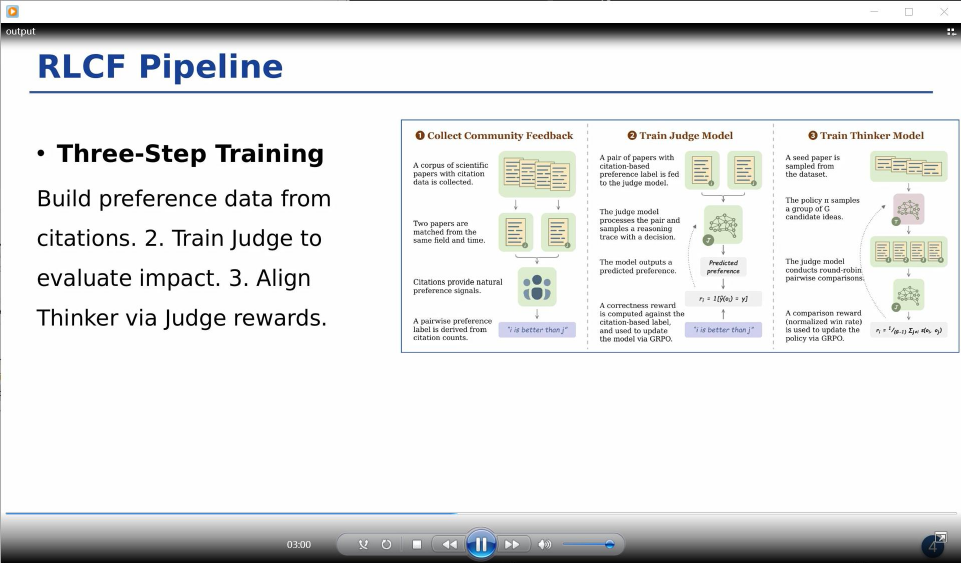}
&
\appcell{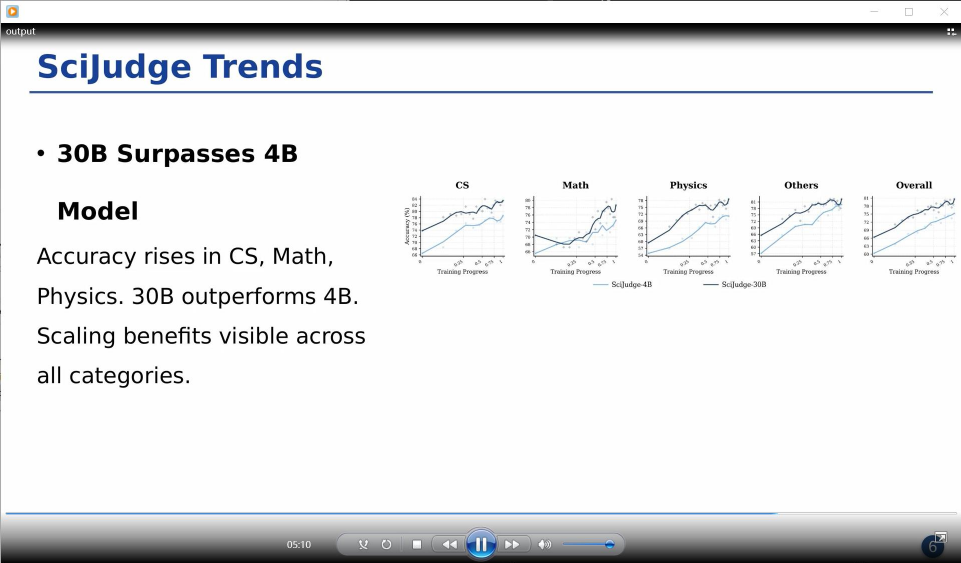}
&
\appcell{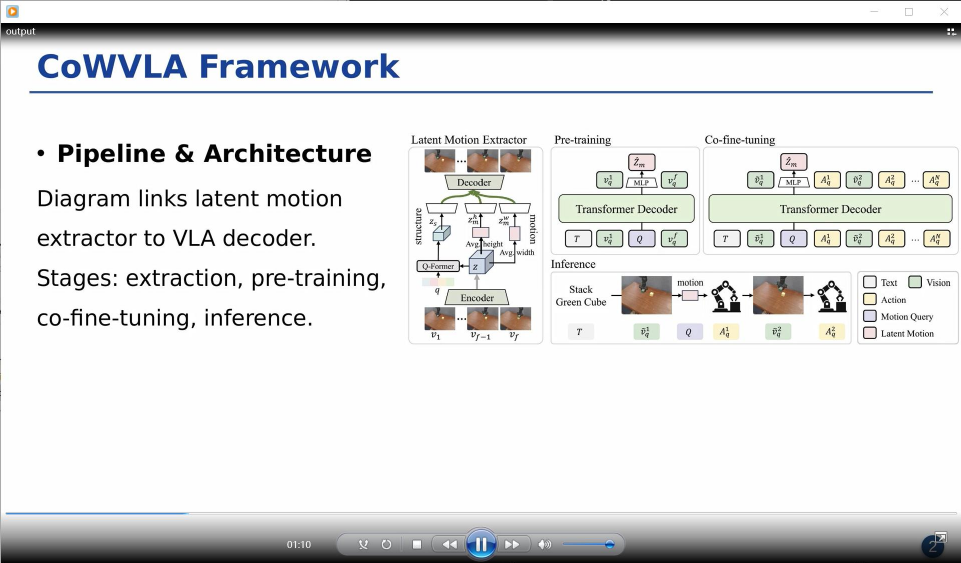}
\\
\noalign{\vskip \appRowGap}

\appcell{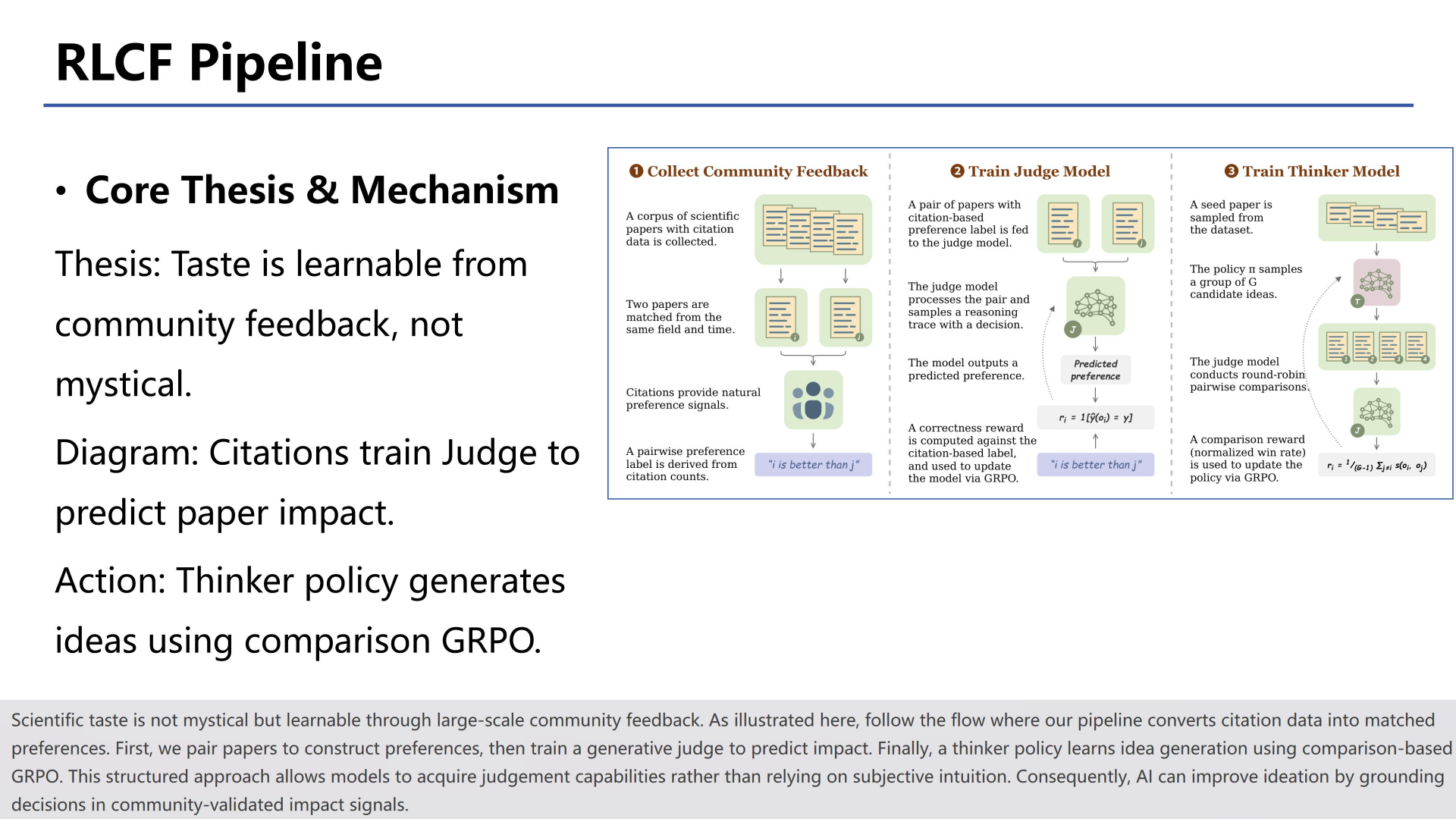}
&
\appcell{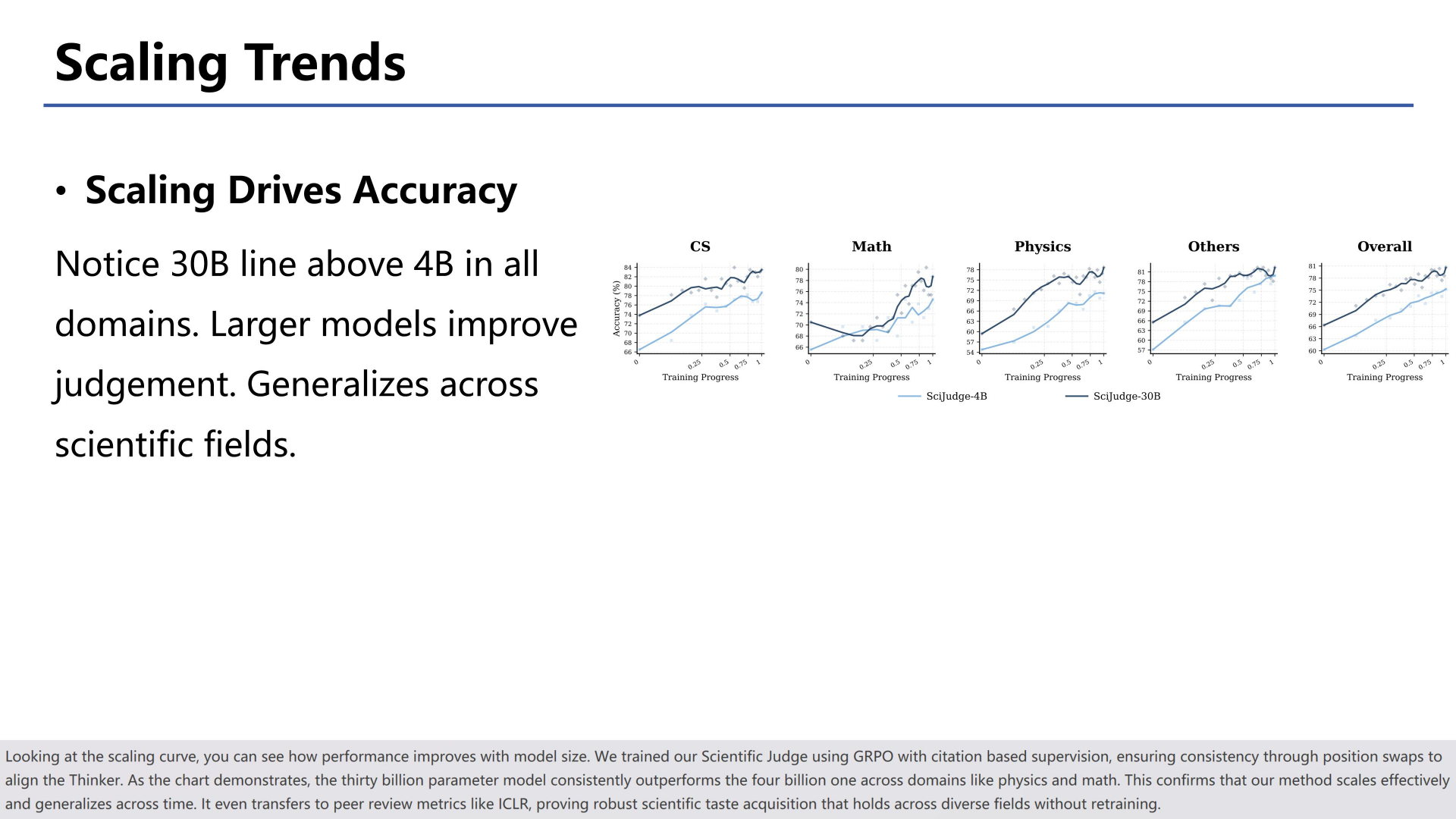}
&
\appcell{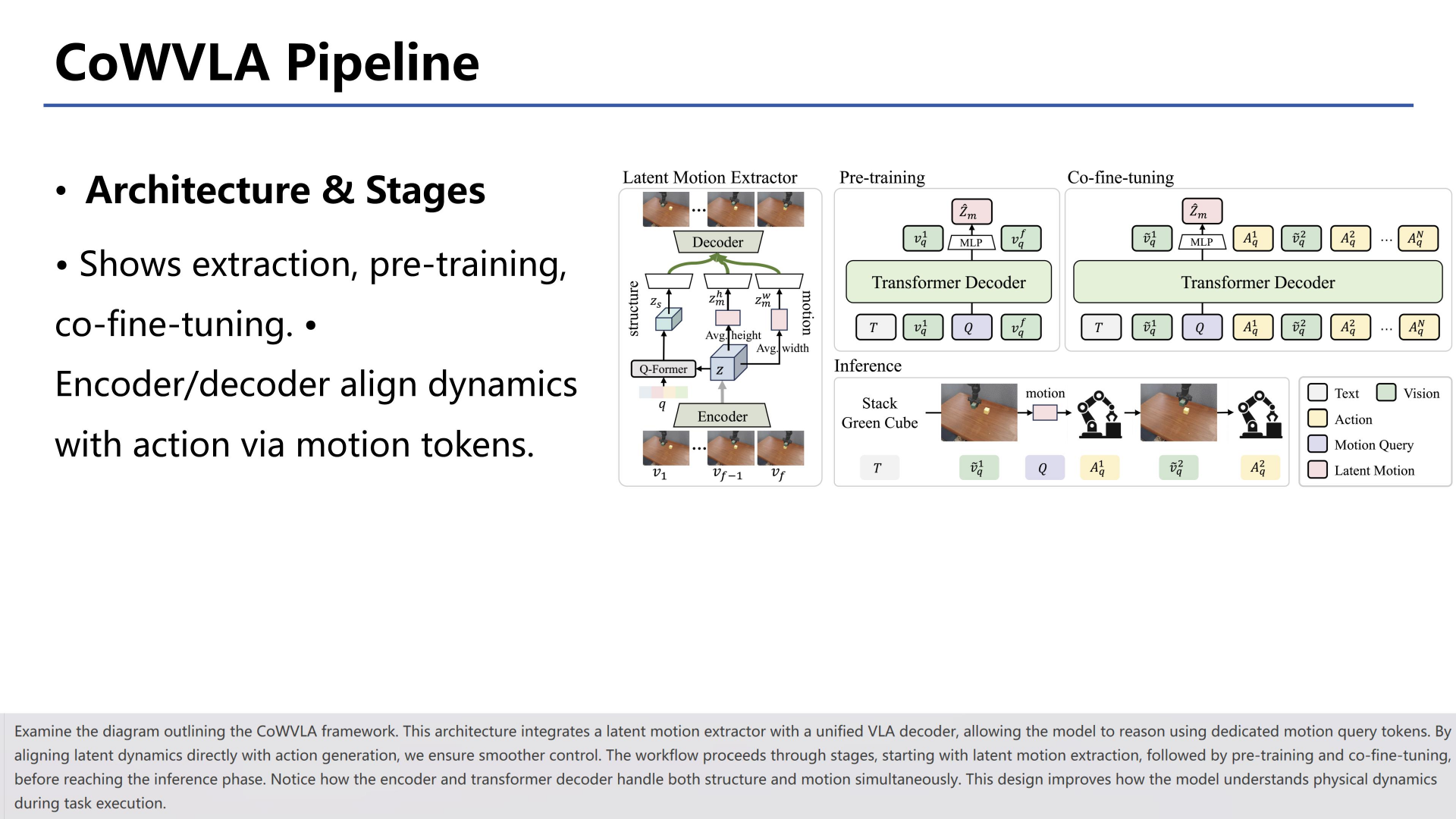}
\\
\noalign{\vskip \appRowGap}

\appcell{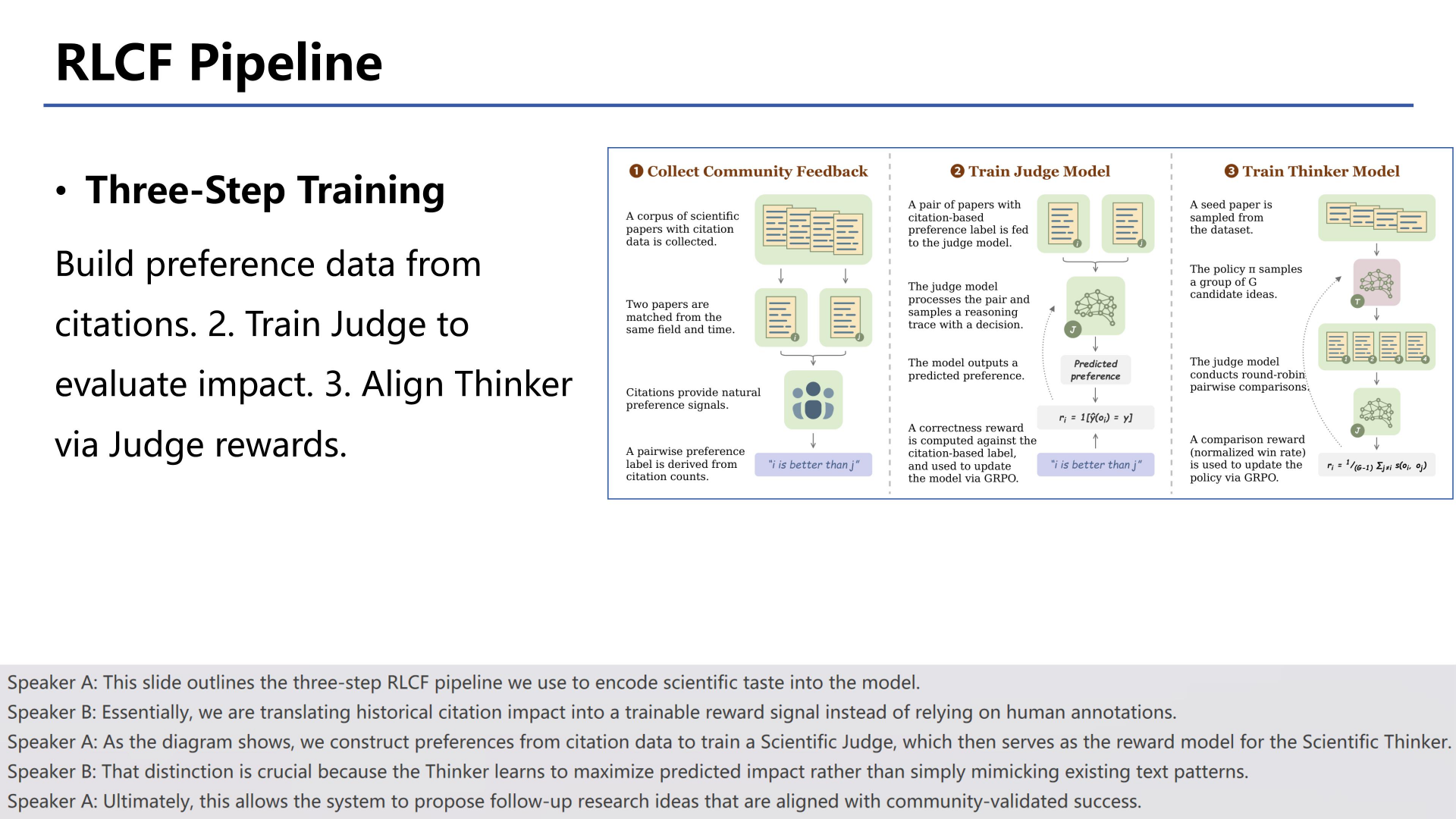}
&
\appcell{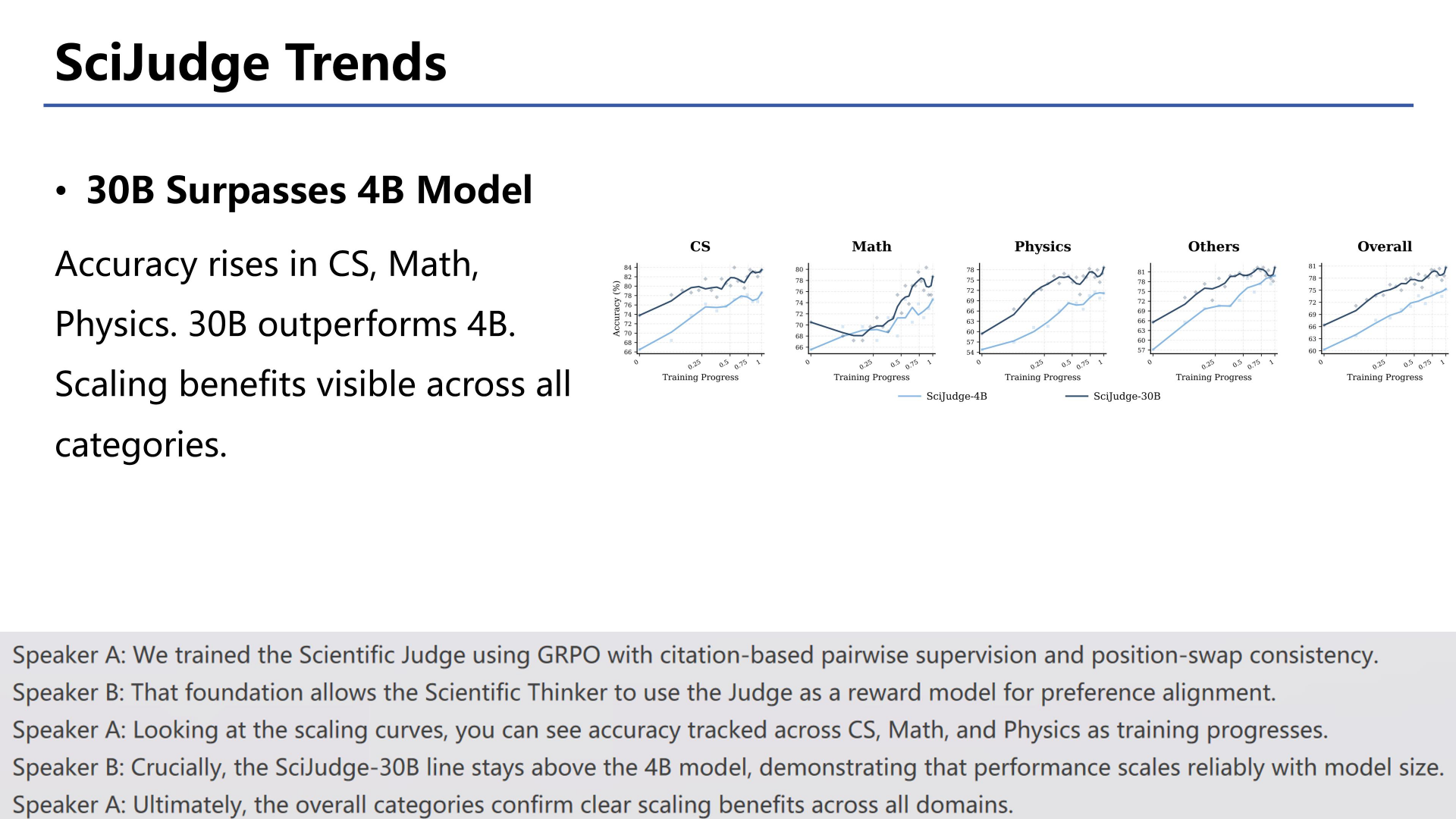}
&
\appcell{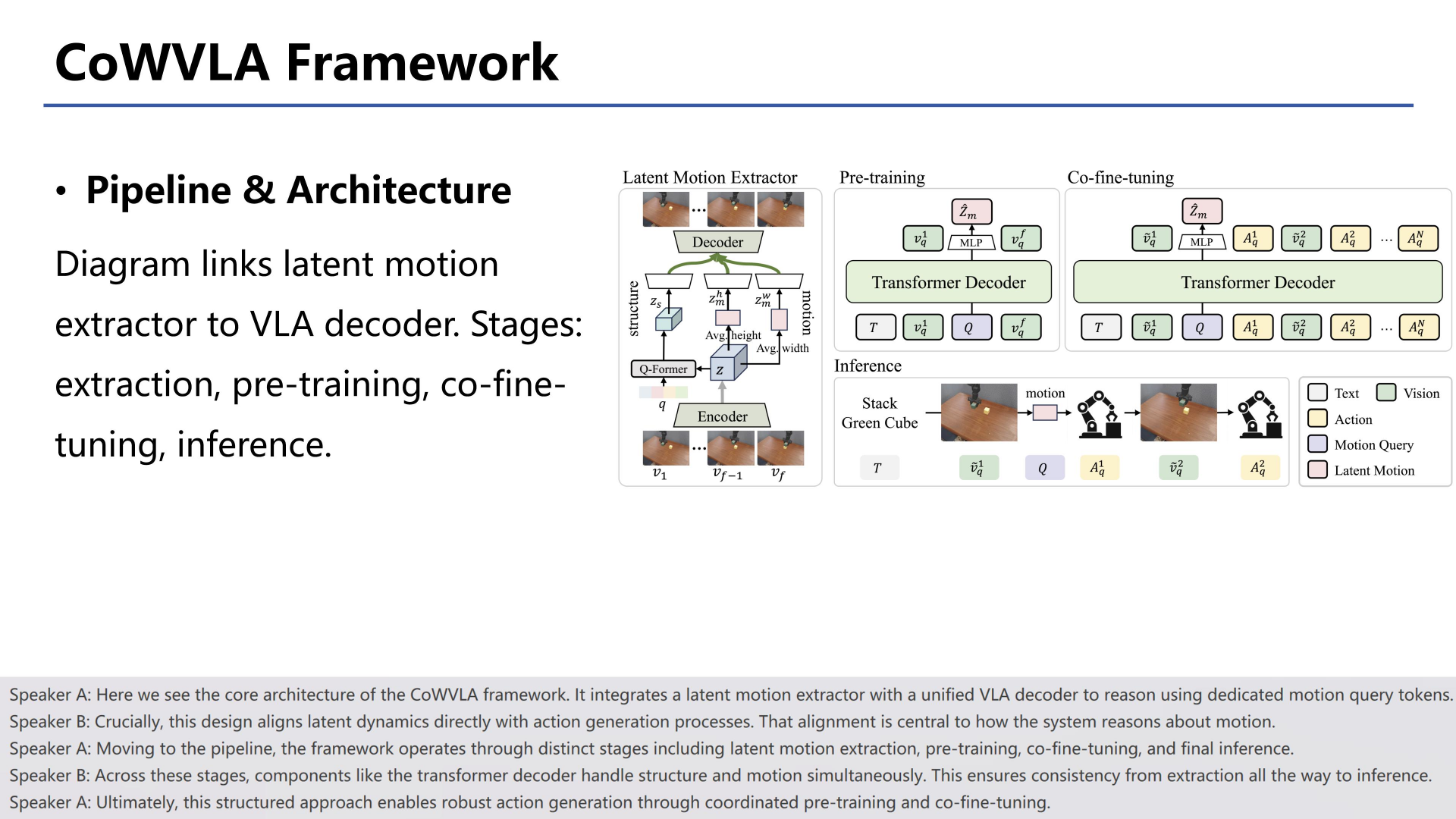}
\\
\noalign{\vskip \appRowGap}

\appcell{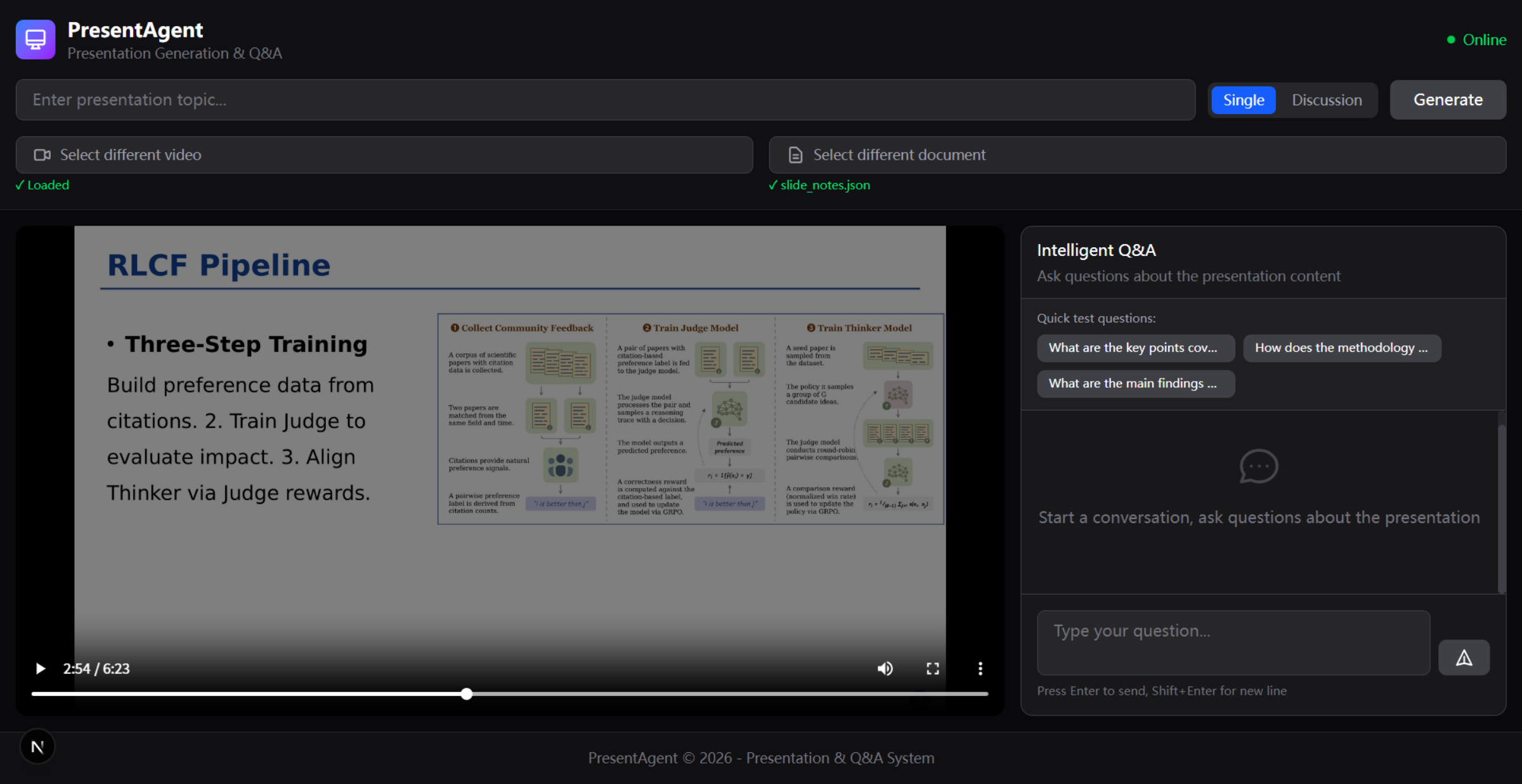}
&
\appcell{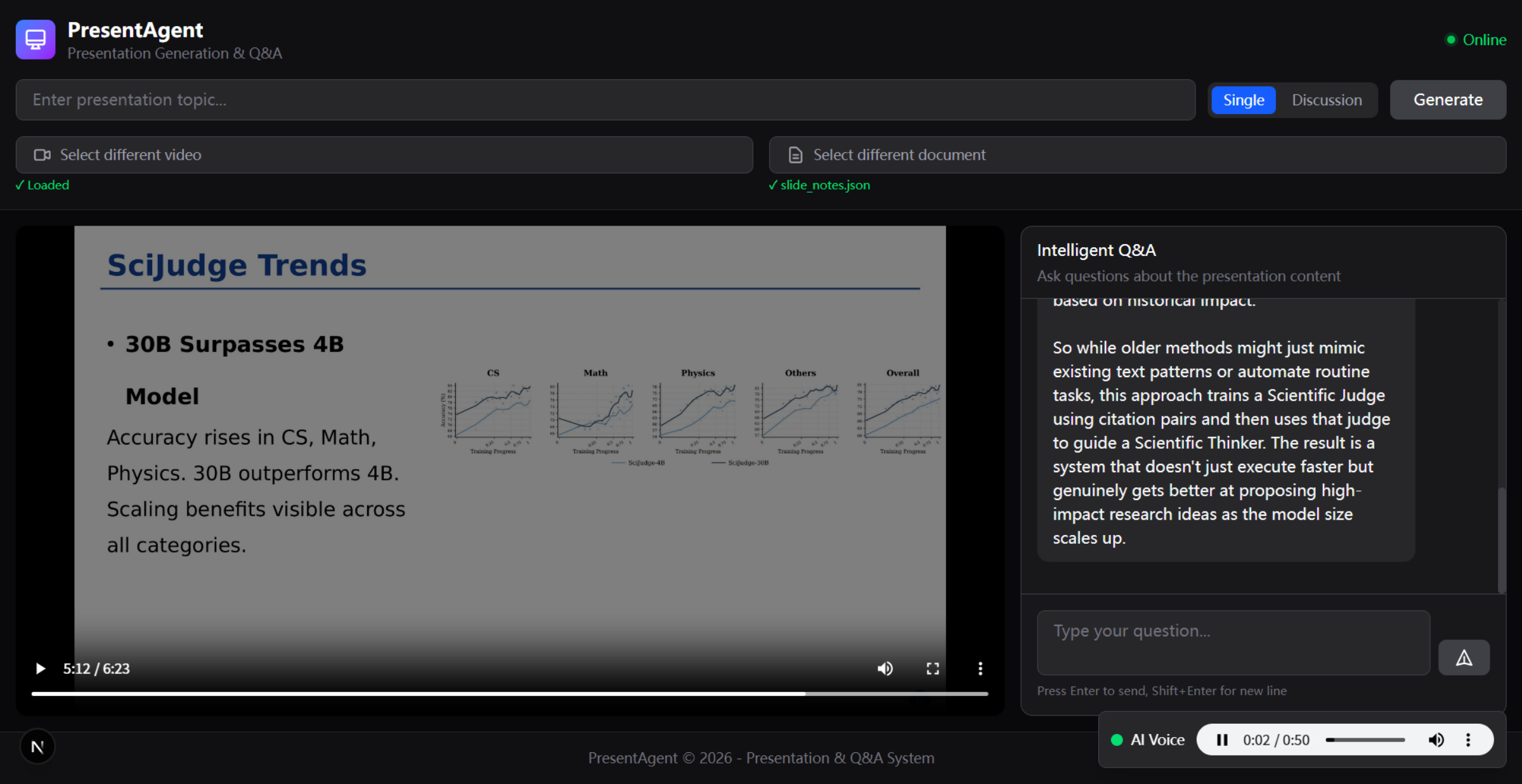}
&
\appcell{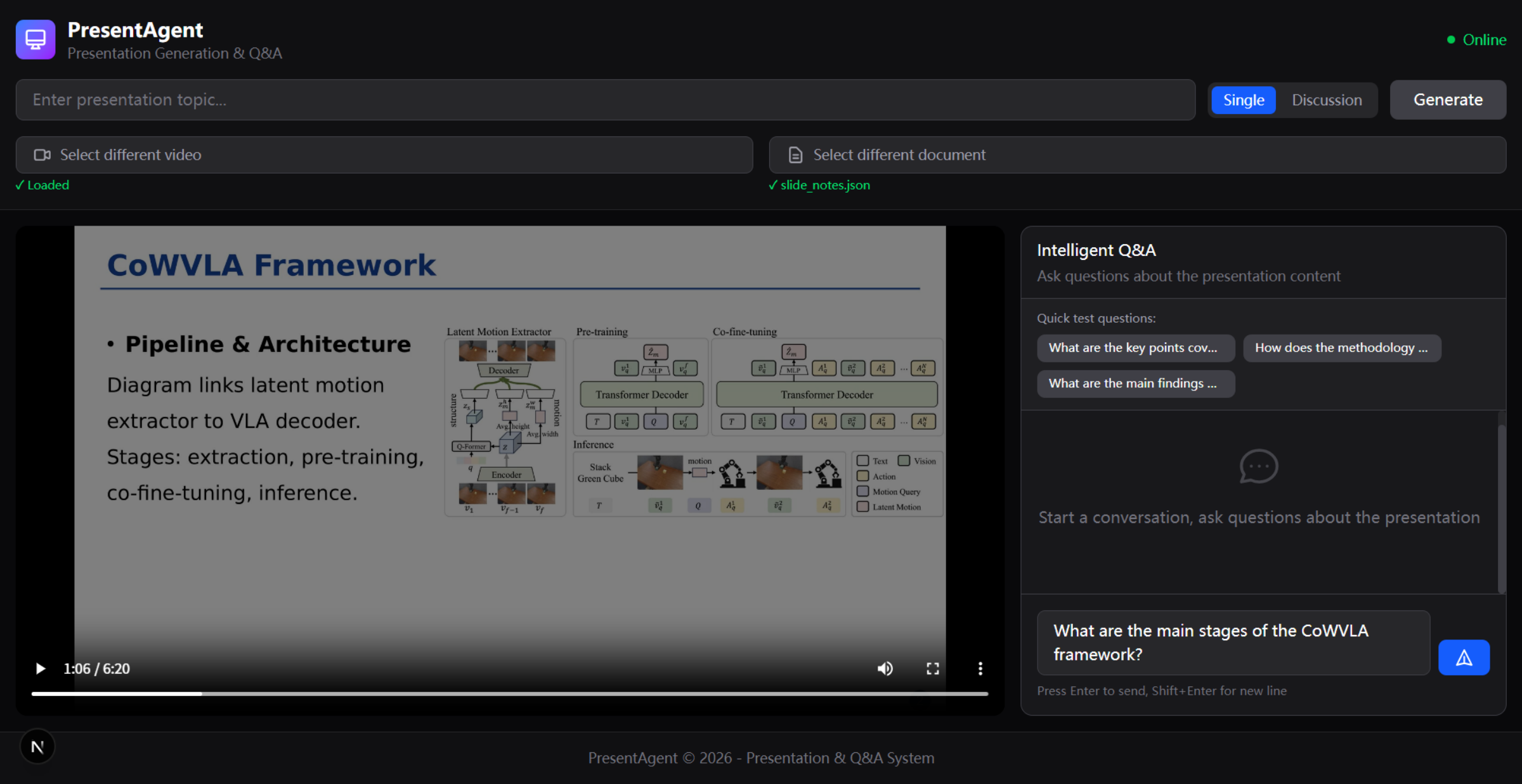}
\end{tabular}

\caption{
Additional qualitative examples of PresentAgent-2, Part 1.
Each column corresponds to one example.
Rows from top to bottom show a representative generated video frame, the generated slide, the generated script, and a representative Interaction Presentation screenshot.
}
\label{fig:appendix_examples_01}
\end{figure*}

\begin{figure*}[h]
\centering
\scriptsize
\setlength{\tabcolsep}{3pt}
\renewcommand{\arraystretch}{1.0}
\setlength{\fboxsep}{0pt}
\setlength{\fboxrule}{0.35pt}

\begin{tabular}{@{}ccc@{}}
\textbf{Example 4}
&
\textbf{Example 5}
&
\textbf{Example 6}
\\[0.25em]

\appcell{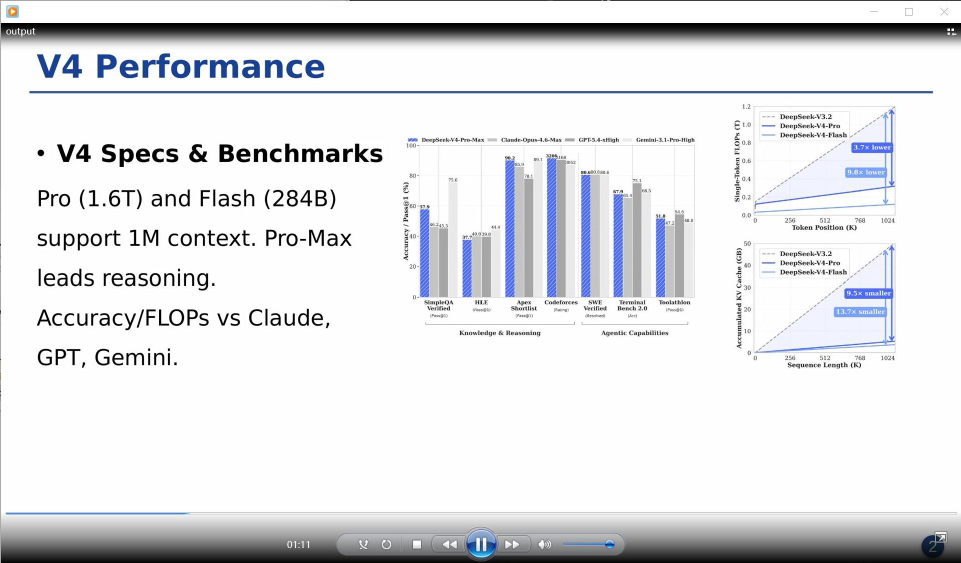}
&
\appcell{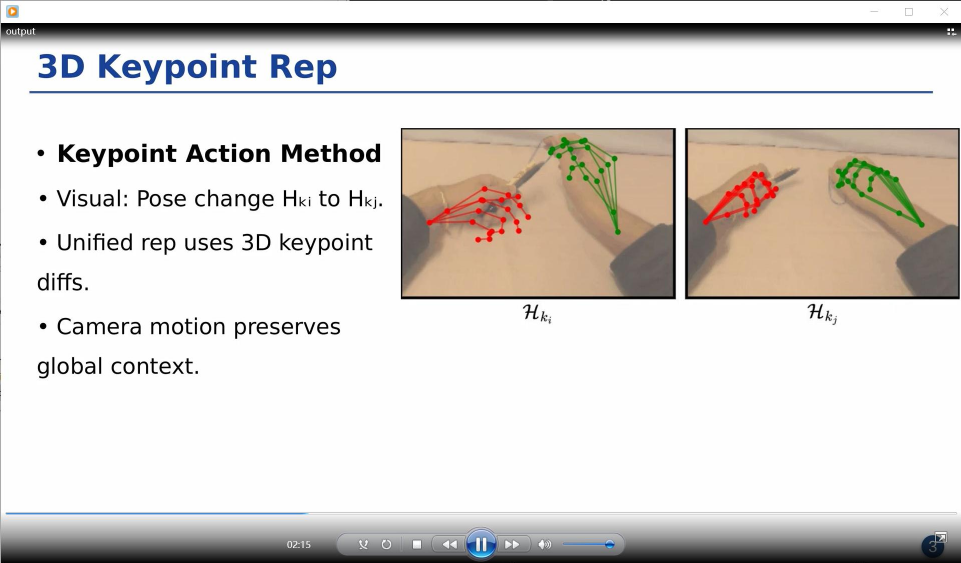}
&
\appcell{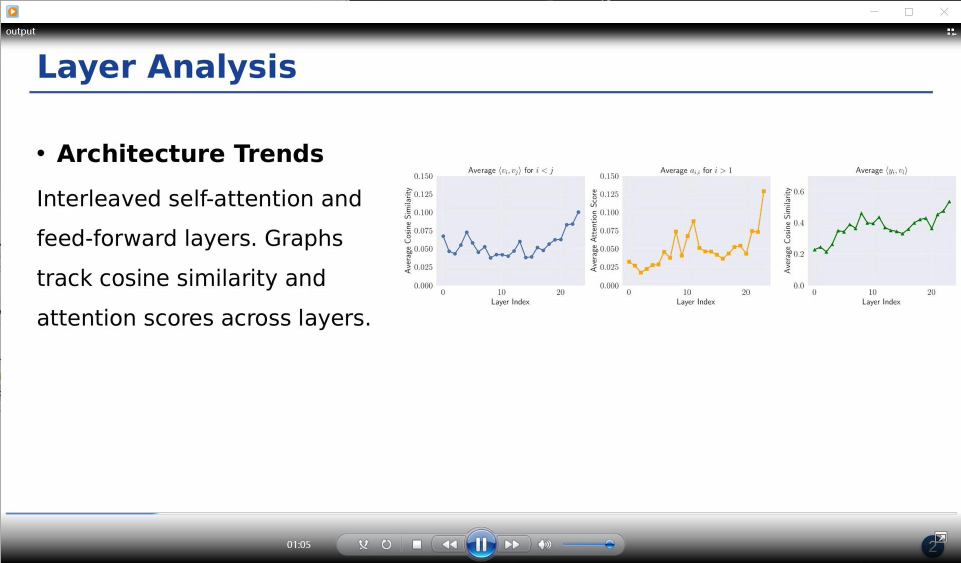}
\\
\noalign{\vskip \appRowGap}

\appcell{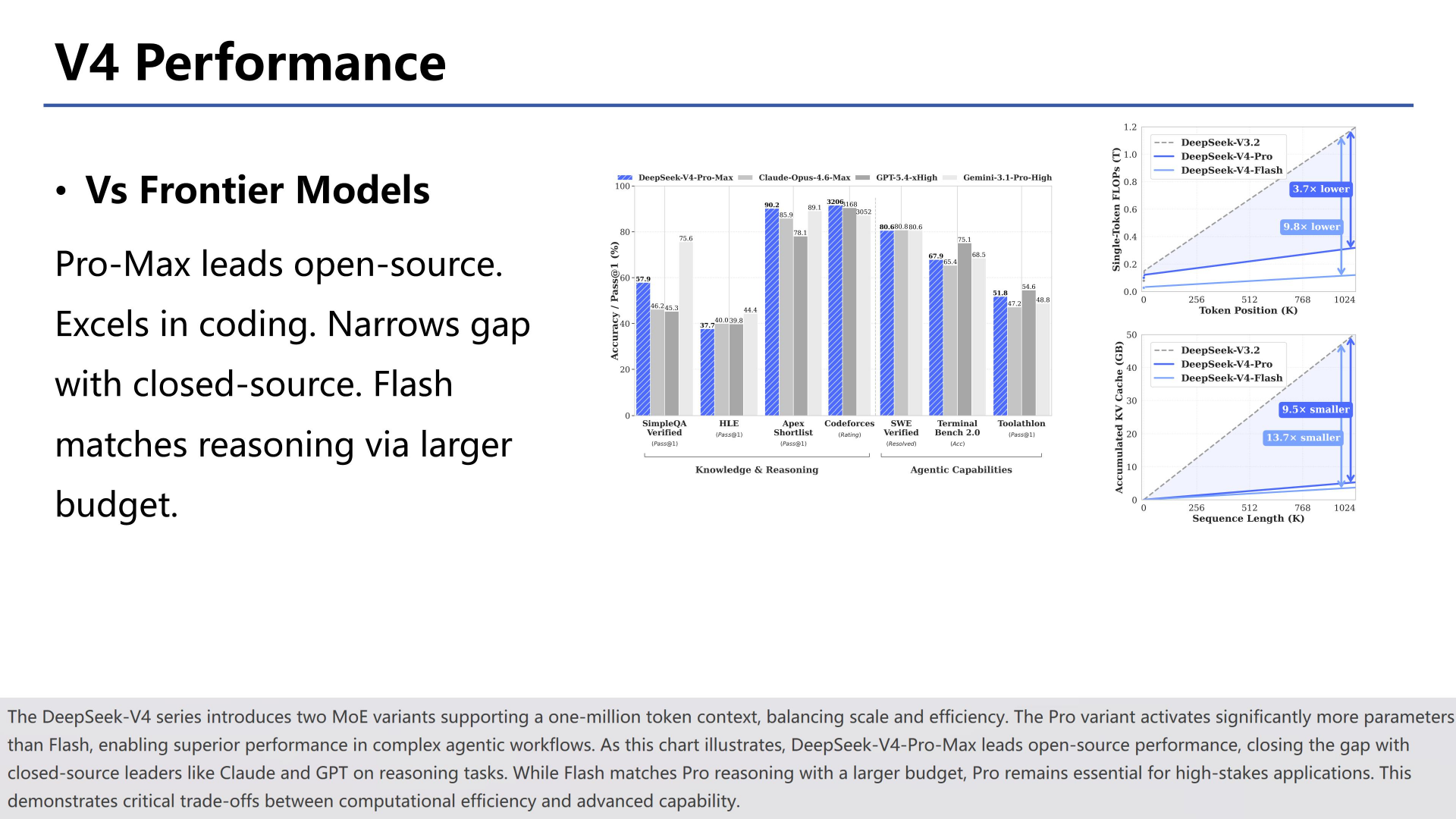}
&
\appcell{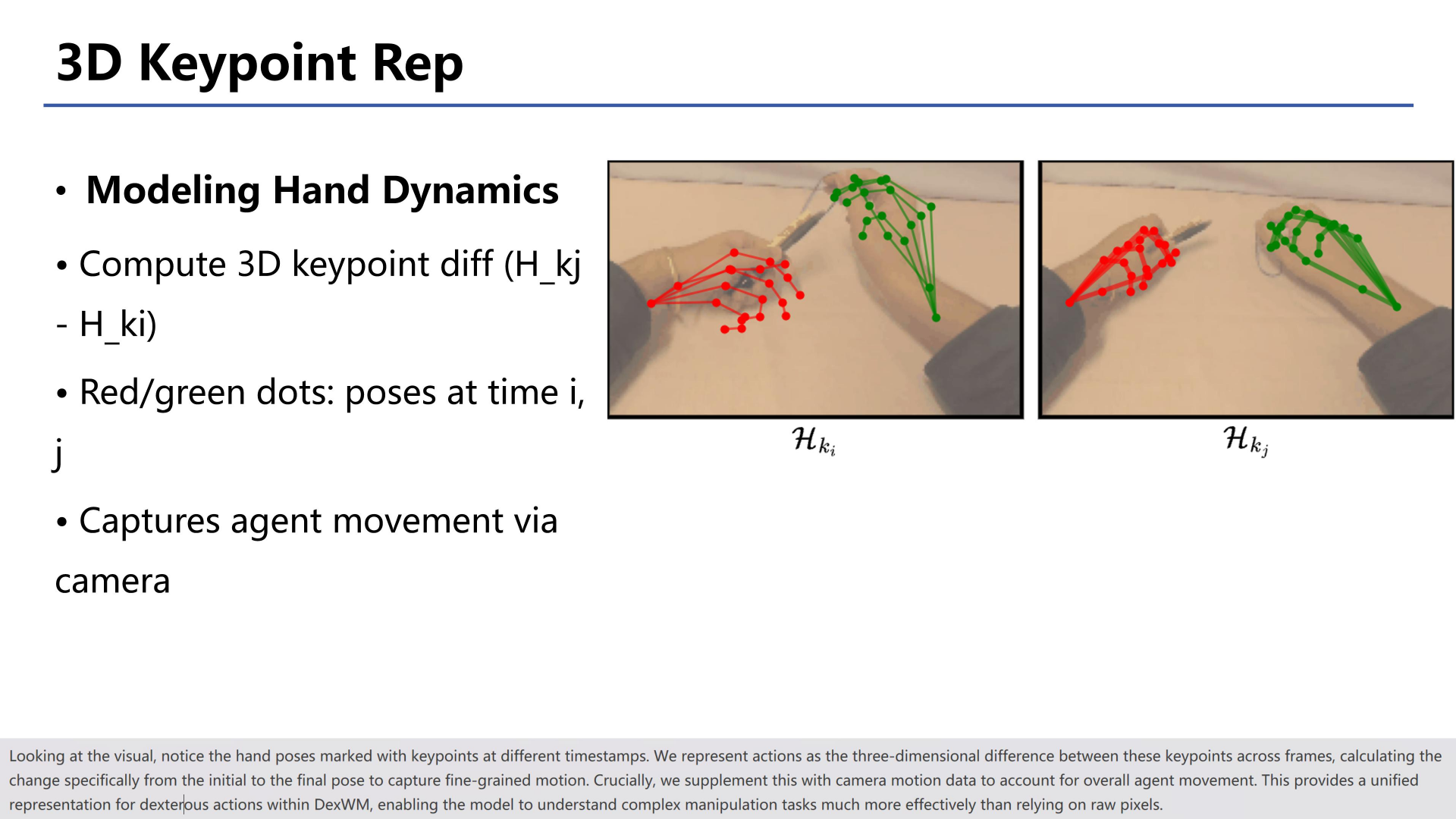}
&
\appcell{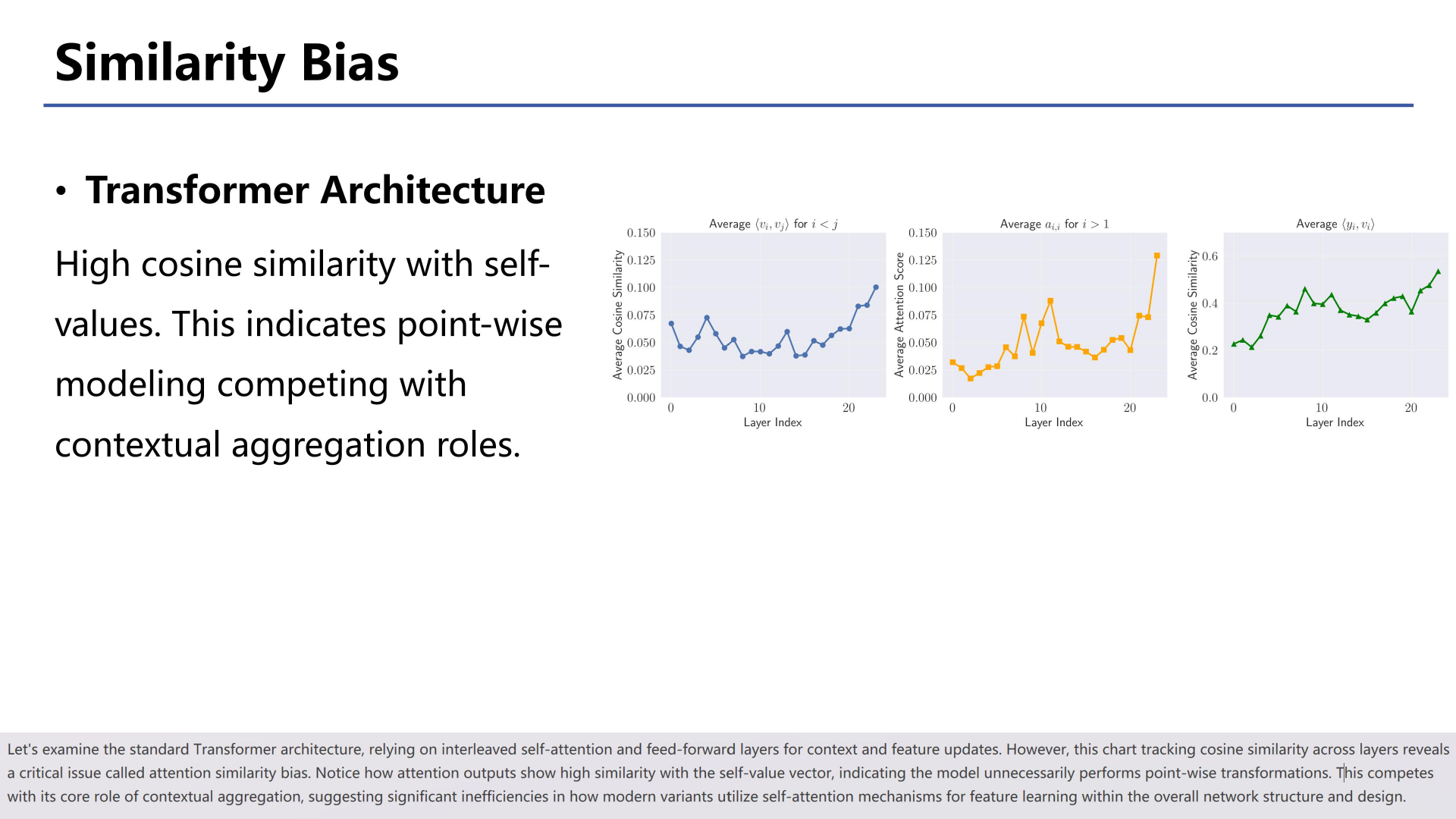}
\\
\noalign{\vskip \appRowGap}

\appcell{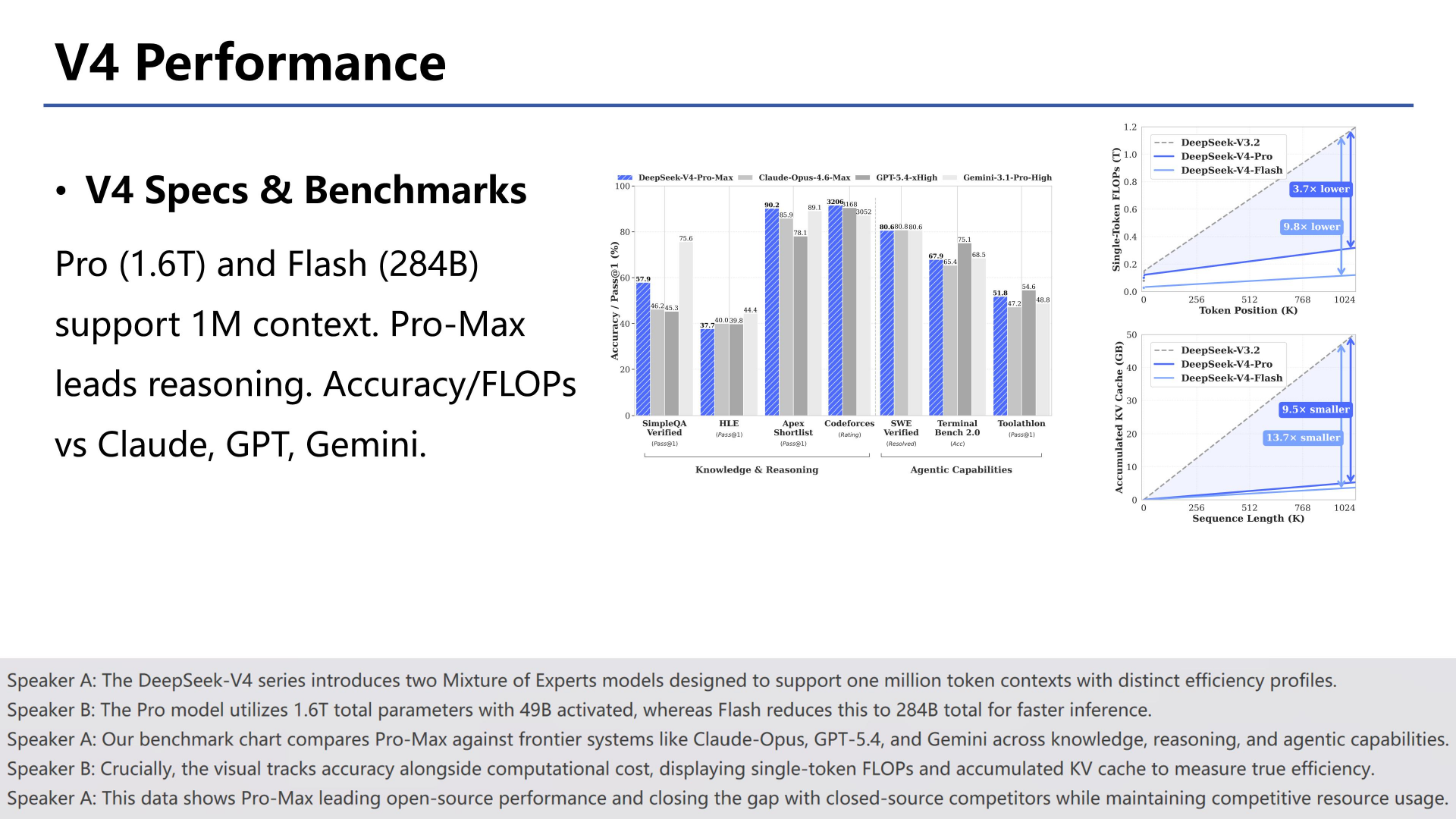}
&
\appcell{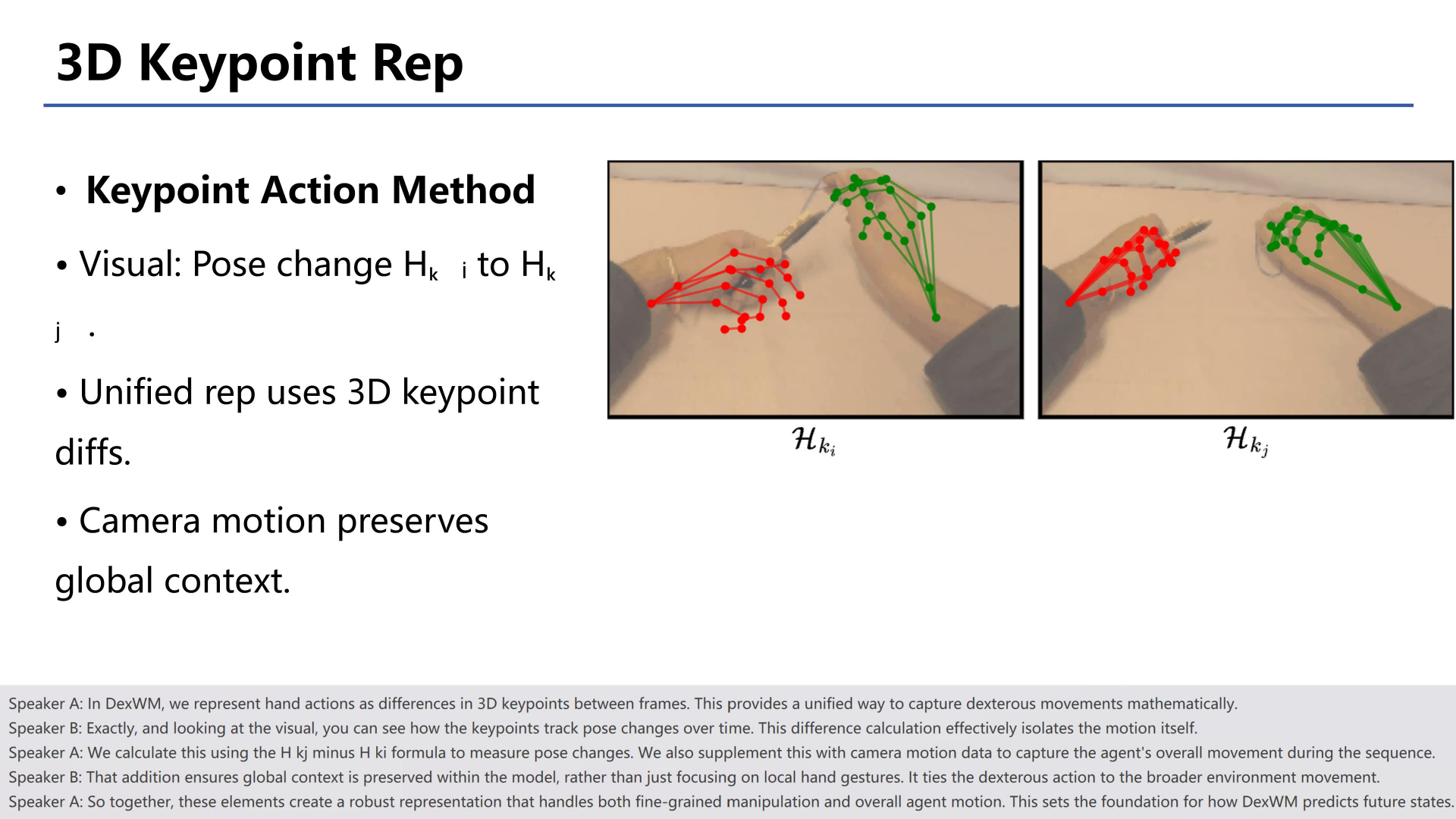}
&
\appcell{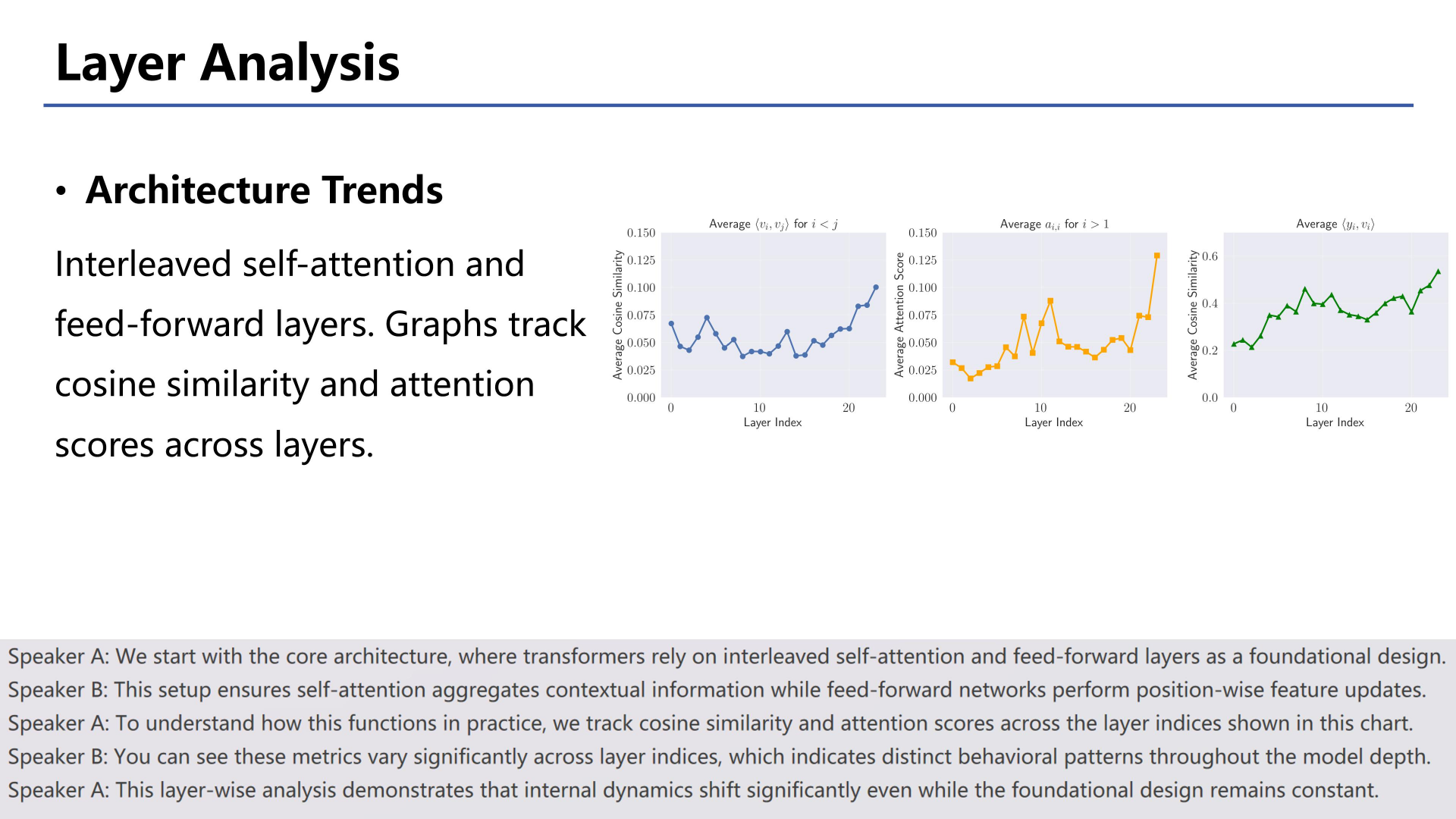}
\\
\noalign{\vskip \appRowGap}

\appcell{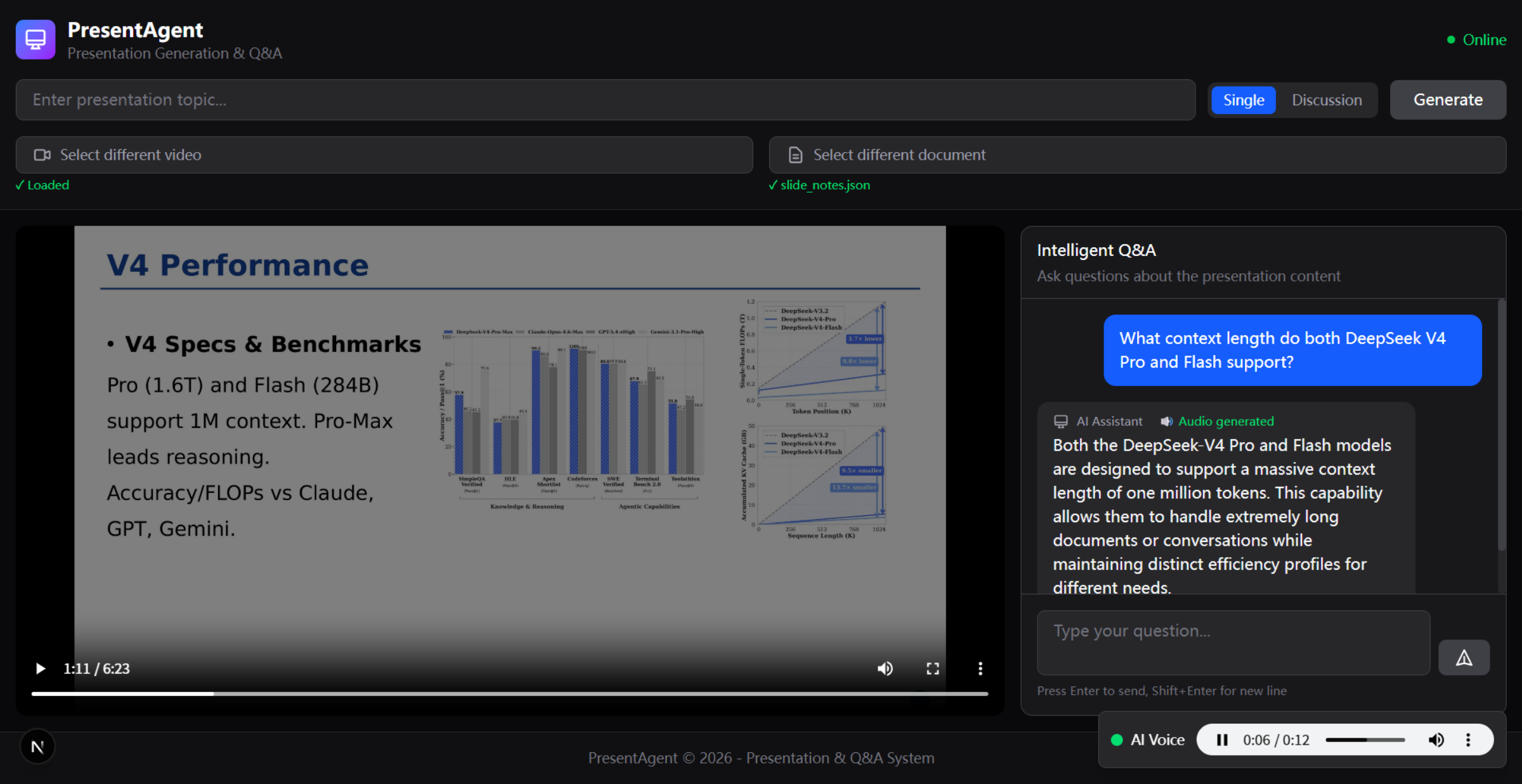}
&
\appcell{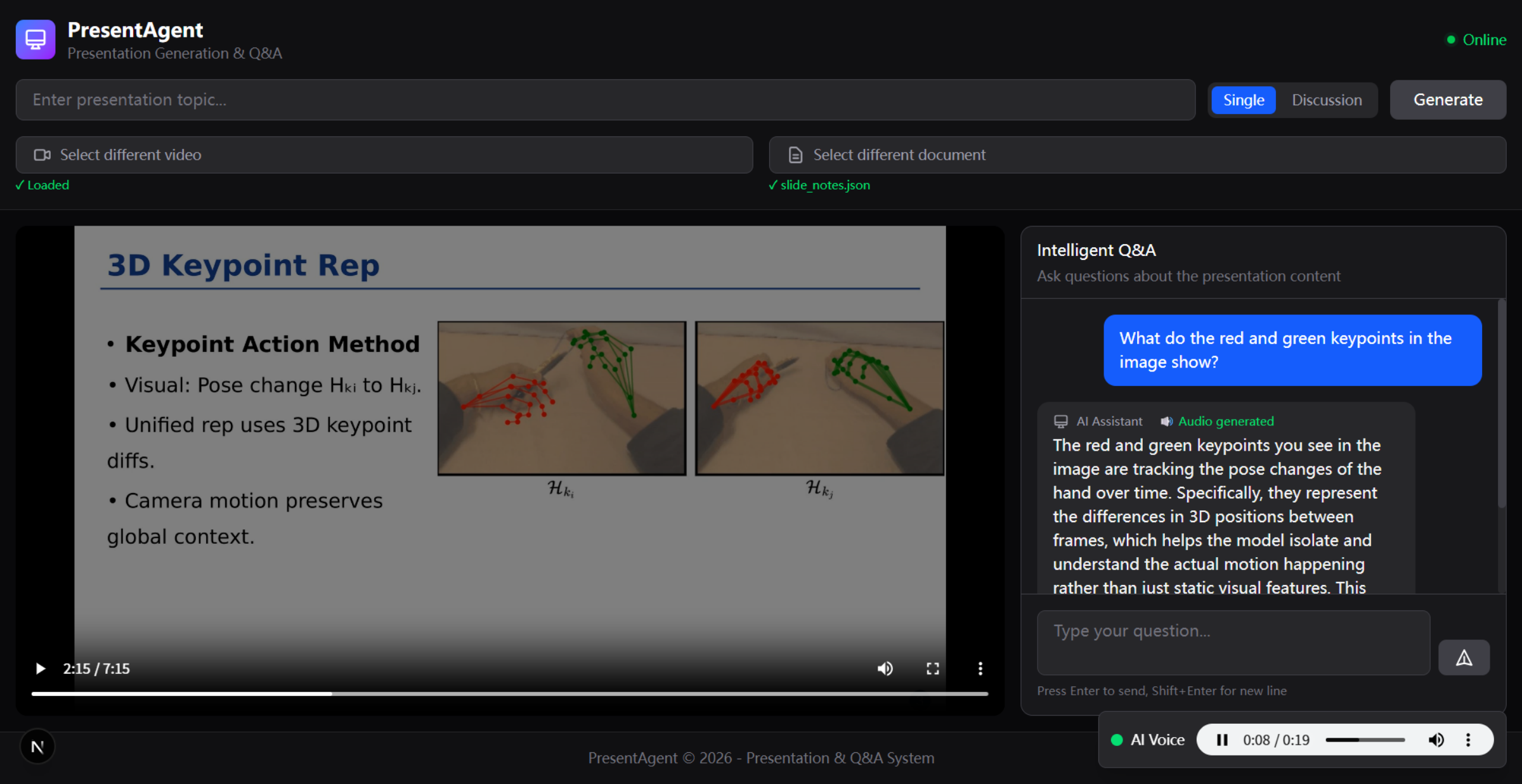}
&
\appcell{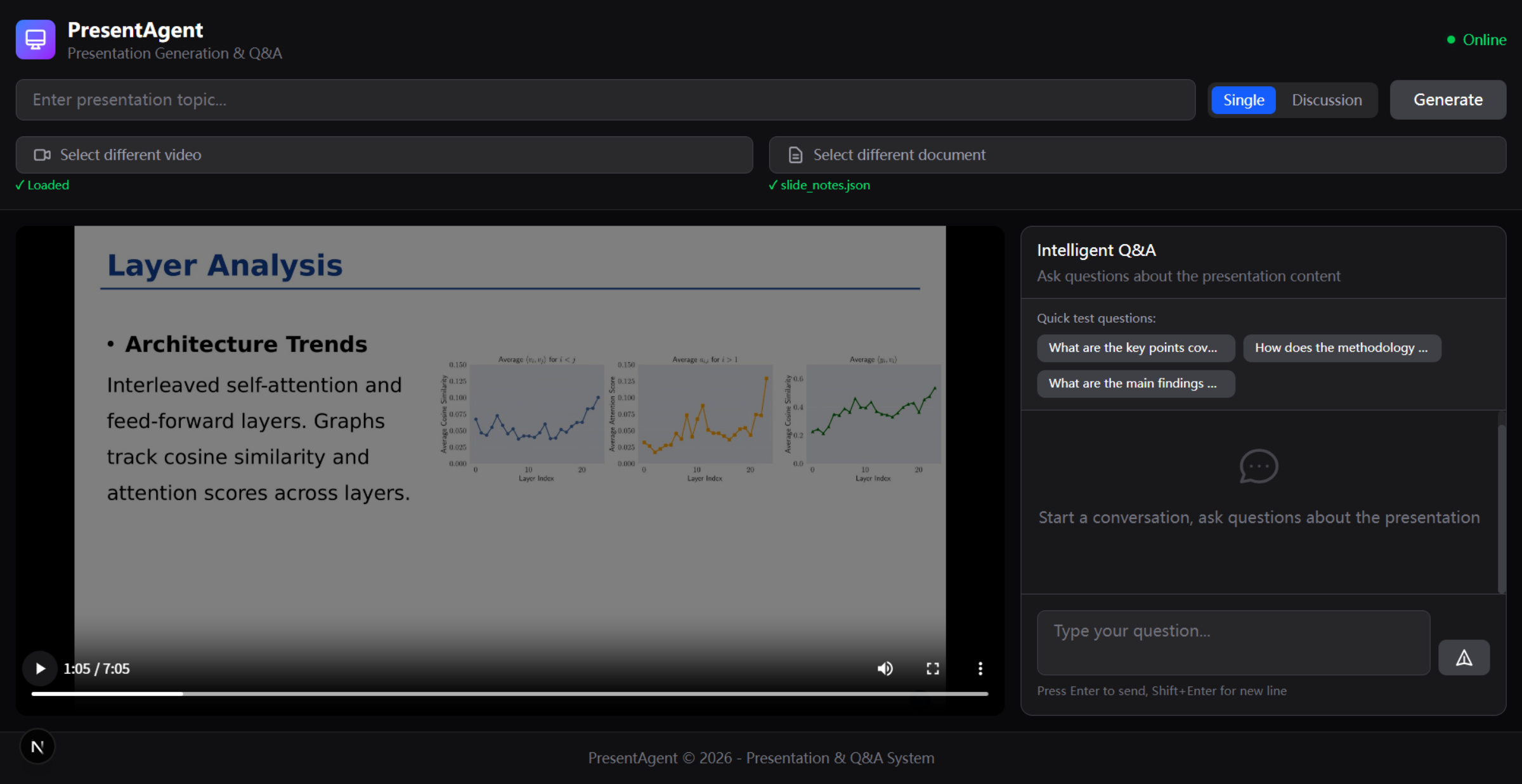}
\end{tabular}

\caption{
Additional qualitative examples of PresentAgent-2, Part 2.
Each column corresponds to one example.
Rows from top to bottom show a representative generated video frame, the generated slide, the generated script, and a representative Interaction Presentation screenshot.
}
\label{fig:appendix_examples_02}
\end{figure*}

\section{Evaluation Prompts and Rubrics}
\label{sec:app_eval_prompts}

\subsection{Objective Quiz Format and Scoring}
\label{sec:app_objective_quiz_scoring}

For objective quiz evaluation, each query--reference video pair is associated with five multiple-choice questions.
Each question contains four options and one correct answer.
The questions are constructed from the corresponding reference presentation video and the expected knowledge points of the user query.
During evaluation, the VLM acts as an audience member and answers the questions using only the generated presentation video and the transcript transcribed from the generated video's audio.

Each quiz question set is stored in a structured format with the following fields:
\begin{itemize}
    \item \texttt{example\_id}: the identifier of the benchmark example;
    \item \texttt{mode}: the presentation mode of the example;
    \item \texttt{questions}: five multiple-choice questions;
    \item \texttt{options}: four answer options for each question;
    \item \texttt{correct\_answer}: the annotated answer key;
    \item \texttt{expected\_knowledge\_point}: the reference knowledge point that the generated video is expected to communicate.
\end{itemize}

For scoring, the predicted answer for each question is compared with the annotated answer key using exact matching.
Each correct answer receives one point, and each incorrect answer receives zero points.
Since each example contains five questions, the quiz score ranges from 0 to 5.
The reported quiz score is the average score over all examples in the corresponding mode and model.

\subsection{Objective Quiz Answering Prompt}
\label{sec:app_quiz_answering_prompt}

The following prompt is used when the VLM acts as an audience member for objective quiz answering.
The VLM is only allowed to use the generated presentation video and the transcript transcribed from the generated video's audio.

\begin{quote}
\small
You are an audience member watching a generated presentation video.

\textbf{Input:}
User Query: \texttt{\{user\_query\}} \\
Generated Presentation Video: \texttt{\{generated\_video\}} \\
Generated Transcript: \texttt{\{generated\_transcript\}} \\
Quiz Questions: \texttt{\{quiz\_questions\}}

\textbf{Task:}
Answer the multiple-choice questions using only the generated presentation video and the generated transcript.

\textbf{Rules:}
Do not use the reference video.
Do not use the annotated correct answers.
Do not use the expected knowledge points.
Do not rely on external knowledge or assume information that is not clearly communicated in the generated video or transcript.
For each question, select exactly one option: A, B, C, or D.

\textbf{Output format:}
Return a valid JSON object containing the predicted answer for each question.
\end{quote}

\subsection{Subjective Scoring Prompt}
\label{sec:app_subjective_scoring_prompt}

For subjective evaluation, the VLM judge assigns independent 1--5 scores to the three metrics defined for the corresponding presentation mode.
The judge receives the user query, generated video, reference video, retrieved resources, and the transcript transcribed from the generated video's audio.

\begin{quote}
\small
You are an audience member and evaluator for a generated presentation video.

\textbf{Input:}
User Query: \texttt{\{user\_query\}} \\
Generated Presentation Video: \texttt{\{generated\_video\}} \\
Reference Presentation Video: \texttt{\{reference\_video\}} \\
Retrieved Resources: \texttt{\{retrieved\_resources\}} \\
Generated Transcript: \texttt{\{generated\_transcript\}} \\
Presentation Mode: \texttt{\{mode\}}

\textbf{Task:}
Evaluate the generated presentation video according to the three metrics defined for the given presentation mode.
Assign an independent score from 1 to 5 for each metric.

\textbf{Scoring scale:}
1 indicates poor performance.
3 indicates acceptable but incomplete performance.
5 indicates strong performance that satisfies the metric definition.

\textbf{Rules:}
Base your judgment on the generated video, transcript, reference video, retrieved resources, and user query.
Do not assign a high score only because the video is visually polished.
The score should reflect whether the generated presentation communicates the requested knowledge and satisfies the requirements of the selected presentation mode.

\textbf{Output format:}
Return a valid JSON object containing the metric scores and a brief justification for each score.
\end{quote}

\paragraph{Query Answering.}
Does the generated video directly answer the user query, cover the key concepts needed to understand the topic, and avoid irrelevant content?

\paragraph{Deep Research Effectiveness.}
Does the generated video effectively use the textual and multimodal resources retrieved through deep research to support the explanation?

\paragraph{Video Delivery Quality.}
Considering the combination of narration, visual examples, and dynamic media, is the video coherent, clear, and easy to follow?

\paragraph{Discussion Effectiveness.}
Does the dialogue format make the content easier to understand than a single-speaker narration by using questions, clarifications, comparisons, or supplementary explanations?

\paragraph{Speaker Role Complementarity.}
Do the different speakers form clear and complementary roles, such as asking questions, explaining, clarifying, or summarizing?

\paragraph{Conversational Delivery.}
Is the discussion natural, engaging, and well-paced, with fluent turn-taking and meaningful question-response flow?

\paragraph{Answer Effectiveness.}
Does the interaction response correctly and directly answer the audience question using information grounded in the presentation context?

\paragraph{Content Comprehensibility.}
Is the interaction response clear, well-structured, and easy to understand, without major ambiguity or confusing explanations?

\paragraph{Interaction Helpfulness.}
Does the interaction response provide useful clarification, connect the answer to the presentation content, and support audience understanding?

\section{Implementation Details}
\label{sec:app_implementation}

We provide additional implementation details for PresentAgent-2. Given a user query and a selected presentation mode, the system first summarizes the query into a focused topic and performs deep research to collect presentation-friendly resources. For each retrieved HTML page, we apply a data-cleaning step to remove boilerplate content, navigation elements, advertisements, and fragmented text, while preserving the main body content. The cleaned page is then evaluated according to content completeness and multimodal richness. Only sources with sufficiently informative textual content and useful multimodal materials, such as images, GIFs, or videos, are retained as candidate resources for presentation generation.

After retrieval and filtering, PresentAgent-2 organizes the collected textual and multimodal resources into a presentation structure. The system plans the slide sequence, generates slide content, writes mode-specific scripts, synthesizes narration audio, and composes the final presentation video. Textual resources are used to support slide titles, bullet points, and explanatory scripts, while visual resources are inserted into the corresponding slide regions to improve visual grounding.

For dynamic media such as GIFs and videos, PresentAgent-2 preserves them as dynamic content during video composition instead of converting them into static screenshots. The dynamic media are placed in the relevant slide regions and synchronized with the generated narration and slide sequence. This allows the final presentation video to retain moving demonstrations, animations, or visual examples when such resources are retrieved during deep research.

The three presentation modes share the same retrieval and presentation generation backbone, but differ in their script and delivery format. Single Presentation uses a single-speaker narration script. Discussion Presentation reformulates the content into a multi-speaker dialogue with complementary speaker roles, such as asking guiding questions, explaining concepts, clarifying details, and summarizing key points. Interaction Presentation supports audience-facing question answering by grounding responses in the generated slides, scripts, retrieved evidence, and presentation context.

\section{Ablation Study}
\label{app:ablation}

We conduct ablation studies to analyze the contribution of key design choices in PresentAgent-2.
All variants use the same backbone model and follow the same evaluation protocol as the main experiments.
To keep the study focused, we separate the ablations into three groups: shared resource ablations, discussion-mode ablations and interaction grounding ablations.
The shared resource ablation evaluates whether multimodal retrieval and dynamic media preservation benefit query-driven presentation generation.
The discussion-mode ablation evaluates whether structured speaker-role assignment is necessary for effective multi-speaker presentations. The interaction grounding ablation evaluates whether grounding interactive Q\&A responses in the full presentation context improves answer quality and conversational coherence.

\paragraph{Shared Resource Ablation.}
We first evaluate the shared resource components used by PresentAgent-2.
\textit{Text-only Retrieval} removes retrieved images, GIFs, and videos from slide and script generation, using only textual resources.
\textit{Static-media} keeps retrieved visual resources but converts GIFs and videos into static frames during video composition.
This ablation tests whether PresentAgent-2 benefits from multimodal evidence and dynamic visual resources beyond text-only retrieval.

\begin{table}[h]
\centering
\caption{
Ablation study of shared resource components.
Text denotes retrieved textual resources; Visual denotes retrieved images, GIFs, and videos; Dynamic denotes preserving GIF/video playback during video composition.
Quiz is the average quiz score on a 0--5 scale, and subjective metrics are rated on a 1--5 scale.
}
\label{tab:shared_resource_ablation}
\scriptsize
\setlength{\tabcolsep}{3pt}
\renewcommand{\arraystretch}{1.08}
\resizebox{\columnwidth}{!}{
\begin{tabular}{lccc|ccccc|ccccc|ccccc}
\toprule
\multirow{2}{*}{\textbf{Variant}}
& \multirow{2}{*}{\textbf{Text}}
& \multirow{2}{*}{\textbf{Visual}}
& \multirow{2}{*}{\textbf{Dynamic}}
& \multicolumn{5}{c|}{\textbf{Single Presentation}}
& \multicolumn{5}{c|}{\textbf{Discussion Presentation}}
& \multicolumn{5}{c}{\textbf{Interaction Presentation}}
\\
\cmidrule(lr){5-9}
\cmidrule(lr){10-14}
\cmidrule(lr){15-19}
&
&
&
& \textbf{Quiz}
& \textbf{QA}
& \textbf{DRE}
& \textbf{VDQ}
& \textbf{Mean}
& \textbf{Quiz}
& \textbf{DE}
& \textbf{SRC}
& \textbf{CD}
& \textbf{Mean}
& \textbf{Quiz}
& \textbf{AE}
& \textbf{CC}
& \textbf{IH}
& \textbf{Mean}\\
\midrule

Text-only Retrieval
& \checkmark
& $\times$
& $\times$
& 4.50 & 4.20 & 3.95 & 4.05 & 4.07
& 4.48 & 4.10 & 3.82 & 4.05 & 3.99 
& 4.60 & 4.55 & 4.10 & 4.19 & 4.28
\\

Static-media
& \checkmark
& \checkmark
& $\times$
& 4.71 & 4.35 & 4.20 & 4.30 & 4.28
& 4.70 & 4.28 & 4.05 & 4.25 & 4.19
& 4.84 & 4.61 & 4.33 & 4.40 & 4.45 \\

\textbf{Full PresentAgent-2}
& \checkmark
& \checkmark
& \checkmark
& \textbf{4.84} & \textbf{4.50} & \textbf{4.48} & \textbf{4.43} & \textbf{4.47}
& \textbf{4.85} & \textbf{4.43} & \textbf{4.22} & \textbf{4.47} & \textbf{4.37}
& \textbf{4.85} & \textbf{4.65} & \textbf{4.43} & \textbf{4.49} & \textbf{4.52}
\\

\bottomrule
\end{tabular}
}
\end{table}

\paragraph{Mode-specific Ablations.}
We further study two mode-specific mechanisms in PresentAgent-2: role-aware dialogue generation for Discussion Presentation and context grounding for Interaction Presentation.
For Discussion Presentation, the full system assigns complementary roles to different speakers, such as question guidance, concept explanation, detail clarification, and summarization.
To test whether this structured role design is necessary, we introduce \textit{Random Script Splitting}, which removes the role-aware prompting logic, first generates a single-speaker narration script, and then assigns its sentences to two virtual speakers.
This ablation examines whether the discussion format benefits from explicit speaker-role complementarity rather than merely splitting a monologue into multiple voices.
For Interaction Presentation, the full system generates audience-oriented responses grounded in the complete presentation context, including structured slides, speaker scripts, and retrieved evidence.
We introduce \textit{Context-Free Interaction}, which removes presentation-context grounding and feeds only the raw audience question into the model.
This ablation examines whether coherent and presentation-consistent interaction relies on comprehensive contextual grounding rather than standalone question answering.
Table~\ref{tab:mode_specific_ablation} reports the results of these mode-specific ablations.

\begin{table}[t]
\centering
\caption{
Mode-specific ablation studies of PresentAgent-2.
The left block evaluates role-aware discussion generation, and the right block evaluates context grounding for interactive presentation.
Quiz is the average quiz score on a 0--5 scale, and subjective scores are rated on a 1--5 scale.
}
\label{tab:mode_specific_ablation}
\scriptsize
\setlength{\tabcolsep}{2.5pt}
\renewcommand{\arraystretch}{1.08}
\resizebox{\columnwidth}{!}{
\begin{tabular}{lc|ccccc|lc|ccccc}
\toprule
\multicolumn{7}{c|}{\textbf{Discussion Presentation}}
& \multicolumn{7}{c}{\textbf{Interaction Presentation}} \\
\cmidrule(lr){1-7}
\cmidrule(lr){8-14}
\textbf{Variant}
& \textbf{Role}
& \textbf{Quiz}
& \textbf{DE}
& \textbf{SRC}
& \textbf{CD}
& \textbf{Mean}
& \textbf{Variant}
& \textbf{Context}
& \textbf{Quiz}
& \textbf{AE}
& \textbf{CC}
& \textbf{IH}
& \textbf{Mean} \\
\midrule

Random Script Splitting
& $\times$
& 4.74
& 4.01
& 3.61
& 3.72
& 3.78
& Context-Free Interaction
& $\times$
& 4.10
& 3.82
& 4.07
& 4.03
& 3.97 \\

\textbf{Full PresentAgent-2}
& \checkmark
& \textbf{4.85}
& \textbf{4.43}
& \textbf{4.22}
& \textbf{4.47}
& \textbf{4.37}
& \textbf{Full PresentAgent-2}
& \checkmark
& \textbf{4.85}
& \textbf{4.65}
& \textbf{4.43}
& \textbf{4.49}
& \textbf{4.52} \\

\bottomrule
\end{tabular}
}
\end{table}

\section{Limitations}
\label{sec:app_limitations}

PresentAgent-2 still has several limitations. 
First, its output quality depends on the availability and reliability of retrieved presentation-friendly sources. 
For queries with limited public multimodal resources or low-quality search results, the generated presentation may contain less informative visual evidence or less comprehensive explanations.

Second, Interaction Presentation relies on the generated slides, scripts, retrieved evidence, and presentation context. 
As a result, errors in upstream retrieval, slide generation, or script generation may propagate to the interaction stage and affect the correctness or helpfulness of grounded answers.

Third, our current benchmark covers 60 query--reference video pairs across Single Presentation, Discussion Presentation, and Interaction Presentation. 
While this setting provides diverse evaluation cases, it does not exhaust all possible presentation domains, audience types, or interaction scenarios. 
Future work can expand the benchmark with more domains, longer presentations, and more fine-grained human evaluation.

\end{document}